\documentclass[lettersize,journal]{IEEEtran}
\usepackage{amsmath,amsfonts}
\usepackage{array}
\usepackage{textcomp}
\usepackage{stfloats}
\usepackage{url}
\usepackage{verbatim}
\usepackage{graphicx}
\usepackage{amssymb}
\usepackage{amsmath}

\usepackage{caption}
\usepackage{cite}
\usepackage[colorlinks,linkcolor=blue,anchorcolor=blue,citecolor=blue]{hyperref}
\usepackage[T1]{fontenc}
\usepackage{booktabs}
\usepackage{multirow}
\usepackage[table,xcdraw]{xcolor}
\usepackage[normalem]{ulem}
\useunder{\uline}{\ul}{}
\usepackage{stfloats}
\usepackage{subcaption}
\usepackage{caption}
\usepackage{graphicx}
\usepackage{hyperref}
\usepackage{array}
\usepackage{hhline}
\usepackage{amsmath}
\usepackage{amsfonts}
\usepackage[utf8]{inputenc}
\usepackage{stfloats}
\usepackage{adjustbox}

\usepackage{array}
\usepackage{enumitem}
\usepackage{units}
\usepackage{makecell}
\usepackage{algorithm}
\usepackage{algpseudocode}

\def\BibTeX{{\rm B\kern-.05em{\sc i\kern-.025em b}\kern-.08em
    T\kern-.1667em\lower.7ex\hbox{E}\kern-.125emX}}
\usepackage{balance}

\begin{document}
\title{Bayesian Fully-Connected Tensor Network for Hyperspectral-Multispectral Image Fusion}
\author{
    Linsong Shan,
    Zecan Yang,
    Laurence T. Yang, ~\IEEEmembership{Fellow,~IEEE}, 
    Changlong Li,
    Honglu Zhao,
    Xin Nie
    \thanks{\textbf{Linsong Shan}, \textbf{Changlong Li} and \textbf{Honglu Zhao} are with the School of Computer Science and Technology, Huazhong University of Science and Technology, Wuhan, 430074, China. 
    \textbf{Laurence T. Yang} is with the School of Computer Science and Technology, Huazhong University of Science and Technology, Wuhan 430074, China, also with the School of Computer Science and Artificial Intelligence, Zhengzhou University, Zhengzhou 450001, China, and also with the Department of Computer Science, St. Francis Xavier University, Antigonish, NS B2G 2W5, Canada.
    \textbf{Zecan Yang} and \textbf{Xin Nie} are with the School of Computer Science and Artificial Intelligence, Zhengzhou University, Zhengzhou 450001, China. 
    E-mail: \textit{\{shanlinsong6, ltyang, zecanyang, aria.hl.zhao\}@gmail.com, changlongli@hust.edu.cn, niexflhx@foxmail.com};

}}

\markboth{}%
{Shell \MakeLowercase{\textit{et al.}}: Bare Demo of IEEEtran.cls for IEEE Journals}
\maketitle

\begin{abstract}
Tensor decomposition is a powerful tool for data analysis and has been extensively employed in the field of hyperspectral-multispectral image fusion~(HMF). Existing tensor decomposition-based fusion methods typically rely on disruptive data vectorization/reshaping or impose rigid constraints on the arrangement of factor tensors, hindering the preservation of spatial-spectral structures and the modeling of cross-dimensional correlations. Although recent advances utilizing the Fully-Connected Tensor Network~(FCTN) decomposition have partially alleviated these limitations, the process of reorganizing data into higher-order tensors still disrupts the intrinsic spatial-spectral structure. Furthermore, these methods necessitate extensive manual parameter tuning and exhibit limited robustness against noise and spatial degradation. To alleviate these issues, we propose the Bayesian FCTN~(BFCTN) method. Within this probabilistic framework, a hierarchical sparse prior that characterizing the sparsity of physical elements, establishes connections between the factor tensors. This framework explicitly models the intrinsic physical coupling among spatial structures, spectral signatures, and local scene homogeneity. For model learning, we develop a parameter estimation method based on Variational Bayesian inference~(VB) and the Expectation-Maximization~(EM) algorithm, which significantly reduces the need for manual parameter tuning. Extensive experiments demonstrate that BFCTN not only achieves state-of-the-art fusion accuracy and strong robustness but also exhibits practical applicability in complex real-world scenarios. \thanks{The source code is available at: \url{https://github.com/LinsongShan/BFCTN}.}

\end{abstract}


\section{Introduction}
\IEEEPARstart{H}{yperspectral} imaging provides a wealth of spectral data across a contiguous bands, which is vital for precise material identification and classification in diverse applications such as computer vision~\cite{chen2024hyperspectral}, remote sensing information~\cite{bioucas2013hyperspectral}, environmental monitoring~\cite{tang2023active}, and agriculture~\cite{barbedo2023review}. However, the acquisition of hyperspectral images~(HSI) is constrained by the limitations of the sensor hardware, and the comprehensive spectral information is frequently obtained at the expense of spatial resolution. In contrast, multispectral imaging~(MSI) captures fewer spectral bands but generally provides higher spatial resolution. To leverage the advantages of both imaging modalities, the fusion of HSI and MSI to produce high-resolution hyperspectral images~(HR-HSI) with both superior spectral and spatial resolutions has emerged as a central research focus in academia and industry alike~\cite{yokoya2017hyperspectral}.

The fusion of hyperspectral and multispectral images has witnessed considerable advancement, driven by advances in diverse methodologies.  This research can be classified into three primary approaches: traditional image fusion techniques, deep learning frameworks, and low-rank decomposition strategies.

Traditional image fusion methods provide a basic framework for HMF tasks. Gomez et al.~\cite{gomez2001wavelet} demonstrated that wavelet transforms effectively enhance both spatial and spectral resolutions through resolution analysis. These methods are simple to implement and preserve essential image details. Pan-sharpening~\cite{loncan2015hyperspectral} further improves hyperspectral images by integrating high-resolution panchromatic and low-resolution spectral data. However, in complex scenes, multi-scale decomposition alone fails to capture the rich spectral information of hyperspectral data.

In recent years, the rise of deep learning has significantly transformed the landscape of image fusion research, enabling automatic extraction of high-level features. In early supervised learning approaches, Wang et al.~\cite{wang2017deep} pioneered the application of CNNs to HMF, while Han et al.~\cite{han2019multi} achieved multi-scale feature extraction through MSCNN. Xiao et al.~\cite{xiao2021physics} proposed a GAN framework further optimized distribution alignment via adversarial training. Dian et al.~\cite{dian2023spectral,dian2024spectral} extended this direction with model-guided and low-rank deep frameworks for spectral super-resolution, effectively integrating spatial–spectral priors. More recently, Liu et al.~\cite{liu2025low} proposed a novel low-rank Transformer architecture that enhances high-resolution imaging through attention mechanisms. To address the dependency on annotated data in supervised learning, Dian et al.~\cite{dian2023zero} developed a zero-shot learning framework, while the teams of Wang~\cite{wang2024unsupervised} and Li~\cite{li2024model,li2025enhanced} established new paradigms through unsupervised Tucker decomposition and enhanced image prior methods, respectively, eliminating the need for supervised data. In addition, Cao et al.~\cite{cao2024unsupervised,cao2024unsupervised2} proposed hybrid Transformer architectures for blind hyperspectral–multispectral fusion, further improving robustness in unsupervised settings. Despite these advancements significantly improving practical applicability, challenges such as model interpretability~\cite{zhu2017deep} and stable generalization across diverse scenarios remain areas for further research, driving the field toward more robust and transparent solutions.

Low-rank decomposition-based image fusion methods leverage the intrinsic low-rank property of images, employing matrix or tensor decomposition techniques to construct unified models that embed multi-source data into a common low-rank representation, thereby achieving effective HMF. Operating in an unsupervised manner without requiring training data, these approaches demonstrate strong adaptability and broad applicability. In early research, Yokoya et al.~\cite{yokoya2011coupled} proposed the CNMF method which effectively preserves spectral-spatial consistency through non-negative matrix factorization, while Simoes et al.~\cite{simoes2014convex} developed the Hysure algorithm utilizing subspace regularization for smooth alignment. However, conventional matrix decomposition methods face limitations in capturing high-order correlations in complex scenes and exhibit sensitivity to noise. In contrast, tensor decomposition provides more effective representation of high-dimensional data~\cite{xue2021multilayer,xue2022laplacian,xue2024tensor}. As natural tensors, hyperspectral images have motivated numerous tensor-based fusion methods. Notable examples include CSTF~\cite{li2018fusing}, which adopts coupled tensor factorization, and NLSTF~\cite{dian2017hyperspectral,dian2019nonlocal}, which incorporates non-local similarity. A series of groundbreaking works, such as LTTR~\cite{dian2019learning}, LTMR~\cite{dian2019hyperspectral}, and GTNN~\cite{dian2024hyperspectral}, further advance this line of research. More recently, CTDF~\cite{xu2024coupled} introduces a fourth-order spatial decomposition scheme. Furthermore, JSSO~\cite{long2024joint} achieves significant improvements in spatial detail reconstruction via joint spatial-spectral optimization. BGS-GTF~\cite{wang2025generalized}, establishes a more generalizable tensor fusion framework capable of handling diverse blurring scenarios.

HMF tasks face challenges such as noise interference, low resolution, and complex spectral–spatial correlations. Bayesian methods show strong potential by naturally incorporating prior knowledge~\cite{yang2025bayesian}, handling noise and missing data, and adaptively adjusting parameters. When combined with tensor decomposition, they can model complex high-dimensional structures with solid theoretical foundations~\cite{zhao2015bayesian}. As a result, Bayesian tensor decomposition has proven effective for improving HMF performance. Representative approaches include Bayesian CP~\cite{ye2023bayesian}, t-SVD~\cite{zhou2019bayesian}, and tensor ring (TR) decomposition~\cite{wang2023fubay}. Recently, Zheng et al.~\cite{zheng2021fully} proposed a FCTN decomposition to capture intrinsic correlations and permutation invariance across tensor modes. Compared with CP and Tucker, FCTN better models high-order interactions and overcomes TR/TT sensitivity to mode arrangements. Jin et al.~\cite{jin2022high} further developed a higher-order coupled FCTN for hyperspectral super-resolution, achieving improved resolution and representation. However, its step-up operation distorts the image structure, and the traditional optimization process requires heavy computation and parameter tuning, limiting generalization.

In this paper, we propose a novel Bayesian Fully-Connected Tensor Network for HMF tasks. The main contributions of our work can be summarized as:

\begin{itemize}
    \item 
    This paper introduces the first integration of Bayesian learning with FCTN decomposition in HMF tasks. The employment of hierarchical sparse priors for the construction of a fully connected tensor network between factor tensors is a novel aspect of this model, which allows flexible cross-dimensional tensor relations and thus better captures low-rank structures and data correlations.
        
    \item 
    A parameter estimation method integrating VB and EM is proposed, effectively addressing non-convex optimization challenges in HMF tasks while streamlining parameter selection. This approach systematically mitigates uncertainties induced by manual parameter tuning.
    
    \item 
    Multi-scenario experiments prove that BFCTN achieves state-of-the-art fusion performance across multiple datasets. Its flexible low-rank modeling capability effectively addresses spatial degradation discrepancies and strong noise interference, while maintaining exceptional robustness and applicability in real-world scenarios.
 
\end{itemize}
The remainder of this paper is organized as follows. Section~\ref{sec:perliminaries} introduces the notation and defines the fusion problem. Section~\ref{sec:model} describes the proposed BFCTN with its prior settings and posterior inference. Section~\ref{sec:exp} reports experimental results and analysis. Section~\ref{sec:conclusion} concludes the paper.

\section{Preliminaries and Problem Formulation}
\label{sec:perliminaries}
\subsection{Notations and Preliminaries}
Tensor is an $n$-dimensional array ($n \geq 3$) that serves as a versatile framework for modeling and analyzing multidimensional data.
 In this article, we adopt lowercase letters, bold lowercase letters, bold uppercase letters, and calligraphic letters to represent scalars, vectors, matrices, and tensors, respectively (e.g., $z$, $\mathbf{z}$, $\mathbf{Z}$, $\mathcal{Z}$). 
 Additionally, given an $N$-th order tensor $\mathcal{Z} \in \mathbb{R}^{I_1\times I_2 \times\cdots \times I_N}$, $\mathcal{Z}(i_1,i_2,\dots,i_N)$ or $z_{i_1,i_2,\dots,i_N}$ represents its $(i_1,i_2,\dots,i_N)$-th element, $\mathbf{Z}_{(n)}$ represents its mode-n unfolding, $vec(\cdot )$ represents its vectorization, and the Frobenius norm is represented as $\left \| \mathcal{Z} \right \|_{\mathrm{F}}$.

FCTN decomposition is the cornerstone of the proposed method. In accordance with the findings of \cite{zheng2021fully}, the element-wise form of FCTN decomposition is as follows:

\begin{equation}
    \begin{aligned}
        {\mathcal{Z}}&\left(i_1, \ldots, i_N\right)=\\
        & \sum_{r_{1,2}}^{R_{1,2}} \ldots \sum_{r_{1, N}}^{R_{1, N}} \sum_{r_{2,3}}^{R_{2,3}} \ldots \sum_{r_{2, N}}^{R_{2, N}} \ldots \sum_{r_{N-1, N}}^{R_{N-1, N}} \\
        & \left\{ {\mathcal{T}_{1}}\left(i_1, r_{1,2}, r_{1,3}, \ldots, r_{1, N}\right)\right. \\
        &{\mathcal{T}_{2}}\left(r_{1,2}, i_2, r_{2,3}, \ldots, r_{2, N}\right) \ldots \\
        & {\mathcal{T}_{n}}\left(r_{1, n}, \ldots, r_{n-1, n}, i_n, r_{n, n+1}, \ldots, r_{n, N}\right) \ldots \\
        & \left.{\mathcal{T}_{N}}\left(r_{1, N}, r_{2, N}, \ldots, r_{N-1, N}, i_N\right)\right\},
        \label{FCTN}
    \end{aligned}
\end{equation}
where $\{\mathcal{T}_{n}\}_{n=1}^{N}$ represents the factor tensors, $i_n$ corresponds to the $N$ dimensions of the original tensor, and $(R_{1,2}, \dots, R_{1,N}, \dots, R_{N-1,N})$ represents the FCTN-ranks. It can be seen that FCTN decomposition decomposes an $N$-th order tensor into $N$ $N$-th order tensors, with its simplified representation as $\mathcal{Z} = \mathbb{FCTN}\left(\{\mathcal{T}_{n}\}_{n=1}^{N}\right)$.

Notably, the mode product of tensors plays a significant role in this study. Assuming the $n$-mode product of a tensor $\mathcal{Z}$ and a matrix $\mathbf{P}$ is denoted as $\mathcal{Z} \times_n \mathbf{P}$, the following equation can be derived based on the properties of FCTN:

\begin{equation}
    \begin{aligned}
        \mathcal{Z} \times_n \mathbf{P} \Leftrightarrow \mathbb{FCTN}\left (\mathcal{T}_1, \mathcal{T}_2,\cdots,\mathcal{T}_n\times \mathbf{P},\cdots,\mathcal{T}_N  \right ).
        \label{modeln}
    \end{aligned}
\end{equation}

Furthermore, the composition of FCTN is of considerable significance. Its expression is $\mathbb{FCTN}\left(\{\mathcal{T}_{n}\}_{n=1}^{N}\right) = \mathcal{Z}$. It should be noted that, in certain cases, it is necessary to isolate a specific factor tensor. The composition of $\{\mathcal{T}_{n}\}_{n=1}^{N}$ without $\mathcal{T}_{n}$ defined to be $\mathcal{Z}_{\neq n}$, assuming $\mathcal{Z}_{\neq n} = \mathbb{FCTN}\left ( \{\mathcal{T}_{n}\}_{n=1}^{N}/\mathcal{T}_{n} \right)$, the mode-n unfolding of the tensor $\mathcal{Z}$ can be expressed as

\begin{equation}
    \begin{array}{l}
        \mathbf{Z}_{(n)}=\left(\mathbf{T}_{n}\right)_{(n)}\left(\mathbf{Z}_{\neq n}\right)_{\left[m_{1: N-1} ; n_{1: N-1}\right]},\\
        \label{Eq:X_n}
    \end{array}
\end{equation}
where
\begin{equation*}
    \begin{array}{l}
        m_{i}=\left\{\begin{array}{ll}
            2 i, & \text { if } i<n, \\
            2 i-1, & \text { if } i \geq n,
        \end{array} \text { and } n_{i}=\left\{\begin{array}{ll}
            2 i-1, & \text { if } i<n \\
            2 i, & \text { if } i \geq n,
        \end{array}\right.\right.
    \end{array}
\end{equation*}
$\left(\mathbf{T}_{n}\right)_{(n)}$ represents the mode-$n$ unfolding of $\mathcal{T}_n$, and $\left(\mathbf{Z}_{\neq n}\right)_{\left[m_{1: N-1} ; n_{1: N-1}\right]}$ represents the mode-$n$ unfolding of the synthesis of other factors. This expansion method is derived from the definition of FCTN decomposition~\cite{zheng2021fully}.

\subsection{Problem Formulation}

With regard to the HMF tasks, the observed images, including low-resolution hyperspectral images (LR-HSI) and high-resolution multispectral images (HR-MSI) are regarded as degraded components of the reference images (HR-HSI). The objective of our study is to restore the reference image by decomposing these degraded components and exploring their low-rank structure. 

The reference image is denoted as $\mathcal{Z}\in \mathbb{R}^{W\times H \times S}  $, where $W$ and $H$ represent the image width and height, respectively, and $S$ represents the number of spectral channels. Additionally, the LR-HSI and HR-MSI are denoted as $\mathcal{H}\in \mathbb{R}^{w\times h \times S}$ and $\mathcal{M}\in \mathbb{R}^{W\times H \times s}$.  The degradation relationship between the observed images and the reference image is as follows:

\begin{equation}
    \begin{aligned}
        \mathcal{H} & = \mathcal{Z} \times_1 \mathbf{P}_1 \times_2 \mathbf{P}_2 + \mathcal{N}_h\\
        \mathcal{M} & = \mathcal{Z} \times_3 \mathbf{P}_3 + \mathcal{N}_m,
    \end{aligned}
    \label{Eq:HM}
\end{equation}
where, $\mathbf{P}_1$ and $\mathbf{P}_2$ are the spatial degradation factors corresponding to width and height, respectively, and $\mathbf{P}_3$ is the spectral degradation factor. The $\mathcal{N}_h$ and $\mathcal{N}_m$ represent distinct noise interference components. With the degradation model above, the HMF tasks is expressed as minimizing the following optimization function:

\begin{equation}
    \begin{aligned}
        \min_\mathcal{Z}\left \| \mathcal{H} - \mathcal{Z}\times_1 \mathbf{P}_1 \times_2 \mathbf{P}_2 \right \|^2_{\mathrm{F}}  + \eta\left \| \mathcal{M} - \mathcal{Z}\times_3 \mathbf{P}_3  \right \| ^2_{\mathrm{F}},
    \end{aligned}
    \label{Eq:min}
\end{equation}
where $\eta$ is a hyperparameter employed to balance the influence of $\mathcal{H}$ and $\mathcal{M}$.

The potential lower-rank structure better accentuates the advantages of FCTN decomposition. Therefore, we assume the reference image $\mathcal{Z} \in \mathbb{R}^{W \times H \times S}$ is chunked into overlapping patches of size $I_1 \times I_2 \times I_3$ with overlap size $p$. Since the observed LR-HSI $\mathcal{H} \in \mathbb{R}^{w \times h \times S}$ is obtained by spatially downsampling $\mathcal{Z}$ by a factor of $sf$, it is chunked into patches of size $I'_1 \times I'_2 \times I_3$, where the spatial dimensions satisfy $I_1 / I'_1 = W / w = I_2 / I'_2 = H / h = sf$. Similarly, the observed HR-MSI $\mathcal{M} \in \mathbb{R}^{W \times H \times s}$ is spectrally degraded from $\mathcal{Z}$, and is thus chunked into patches of size $I_1 \times I_2 \times I'_3$, where $I_3 / I'_3 = S / s$. Subsequently, the image patches are classified into $K$ groups in accordance with the columns of their original positions, with $I_4$ representing the number of image patches in this group. As a result, $K$ groups of LR-HSI and HR-MSI image patches are generated, with each group denoted by $\mathcal{H}^{k} \in \mathbb{R}^{I'_1 \times I'_2\times I_3 \times I_4}$ and $\mathcal{M}^{k} \in \mathbb{R}^{I_1 \times I_2\times I'_3 \times I_4}$. The corresponding reference image patch is represented as $\mathcal{Z}^{k} \in \mathbb{R}^{I_1 \times I_2\times I_3 \times I_4}$. Additionally, the relationship between $\mathcal{H}^k$, $\mathcal{M}^k$ and $\mathcal{Z}^k$ still satisfy Eq. \eqref{Eq:HM}. While chunking the observed image into overlapping patches increases computational costs, it facilitates more effective feature extraction. This group-wise strategy avoids explicitly modeling the spatial arrangement of patches, thus preventing the computational need to construct tensors of even higher order.

\begin{figure} [h!]
    \includegraphics[width=\linewidth]{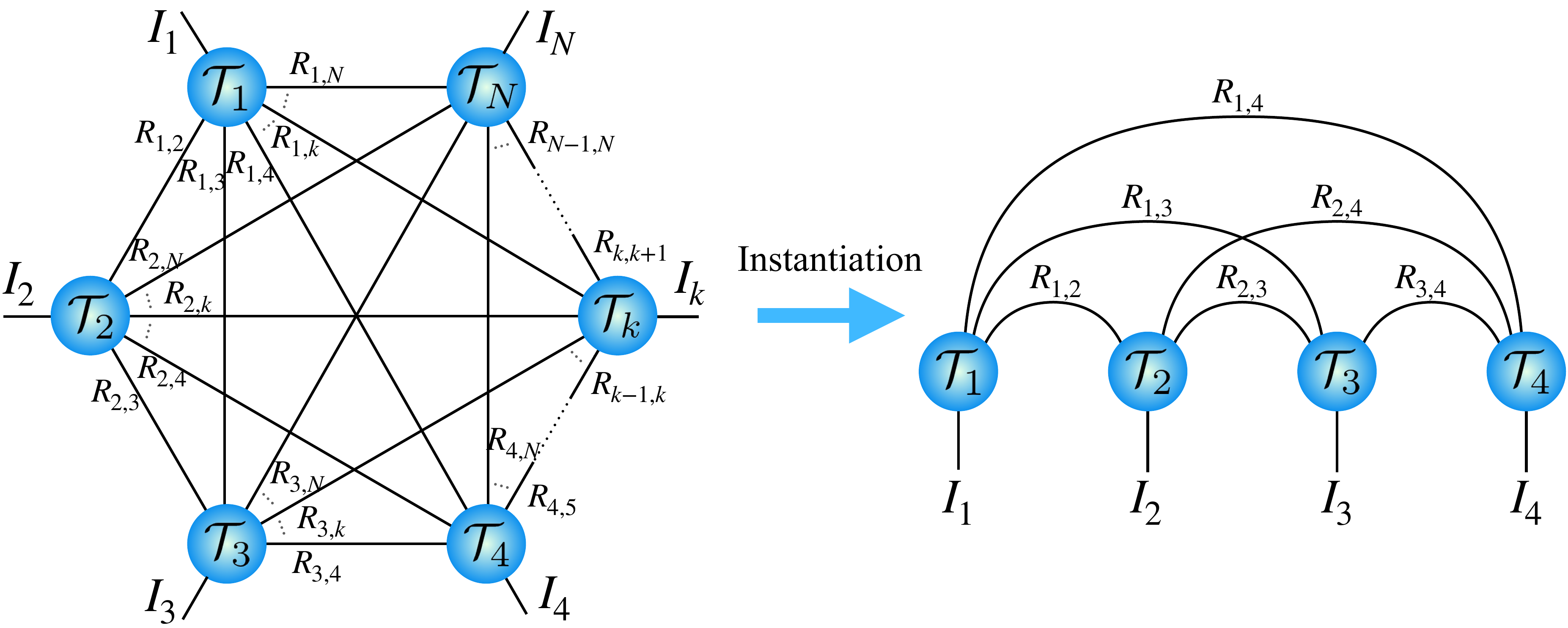}
    \caption{Graphical representation of FCTN decomposition and its 4th-order instantiation.}
    \label{fig:FCTN}
\end{figure}

The standard FCTN models a tensor $\mathcal{Z}$ as the product of multiple factor components across $N$ modes. However, in the HMF task, decomposing the spectral tensor into an excessively high-order structure may disrupt the inherent spatial–spectral correlations, thereby reducing physical interpretability and robustness under complex degradations. Therefore, the number of factor tensors is restricted to four, corresponding to the observed 4th-order tensorized image. Fig.~\ref{fig:FCTN} shows the structure of the FCTN decomposition and its 4th-order graphical representation. In addition, the FCTN decomposition of $\mathcal{Z}^k$ can be denoted as

\begin{equation}
    \begin{aligned}
        &\mathcal{Z}^k = \mathbb{FCTN}\left(  \mathcal{T}^k_1, \mathcal{T}^k_2, \mathcal{T}^k_3, \mathcal{T}^k_4 \right ),\\
    \end{aligned}
    \label{Eq:FCTN_Z}
\end{equation}

where $ \mathcal{T}^k_1 \in \mathbb{R}^{I_1 \times R_{1,2}\times R_{1,3} \times R_{1,4}} $, $ \mathcal{T}^k_2 \in \mathbb{R}^{R_{1,2} \times I_2 \times R_{2,3} \times R_{2,4}} $, $ \mathcal{T}^k_3 \in \mathbb{R}^{R_{1,3} \times R_{2,3}\times I_3 \times R_{3,4}} $, $ \mathcal{T}^k_4 \in \mathbb{R}^{R_{1,4} \times R_{2,4}\times R_{3,4} \times I_4} $ are factor tensors, and $\{R_{n_1,n_2}\}_{1\le n_1<n_2\le4}$ is FCTN ranks. 


Based on Eq.~\eqref{modeln} and Eq.~\eqref{Eq:HM}, the image patches $\mathcal{H}^{k}$ and $\mathcal{M}^{k}$ can be expressed in terms of factor tensors as
\begin{equation}
    \begin{aligned}
        &\mathcal{H}^k = \mathbb{FCTN}\left( \mathcal{T}^k_1 \times_1 \mathbf{P}_1, \mathcal{T}^k_2\times_2 \mathbf{P}_2, \mathcal{T}^k_3 ,\mathcal{T}^k_4\right )\\   
        &\mathcal{M}^k = \mathbb{FCTN}\left( \mathcal{T}^k_1, \mathcal{T}^k_2, \mathcal{T}^k_3\times_3 \mathbf{P}_3,\mathcal{T}^k_4 \right ).
    \end{aligned}
    \label{Eq:FCTN_HM}
\end{equation}
With the above considerations, the optimization function in Eq.~(\ref{Eq:min}) can be rewritten as follows:
\begin{equation}
    \begin{aligned}
        &\min_{\{ \mathcal{T}^{k}_{n}\}^4_{n=1}}\left \| \mathcal{H}_k - \mathbb{FCTN}\left ( \mathcal{T}^{k}_{1}\times_1 \mathbf{P}_1,\mathcal{T}^{k}_{2}\times_2 \mathbf{P}_2,\mathcal{T}^{k}_{3},\mathcal{T}^{k}_{4}\right) \right \|^2_{\mathrm{F}} \\ 
        & \qquad +\eta \left \| \mathcal{M}_k -\mathbb{FCTN}\left ( \mathcal{T}^{k}_{1}, \mathcal{T}^{k}_{2}, \mathcal{T}^{k}_{3} \times_3 \mathbf{P}_3, \mathcal{T}^{k}_{4} \right)  \right \|^2_{\mathrm{F}}.
    \end{aligned}
    \label{Eq:min2}
\end{equation}

\section{Model Description}
\label{sec:model}
This section will present the proposed model in the form of a probabilistic graphical model. The model can be summarised in terms of two principal components: the model priors setting and the model inference.
\vspace{-5mm}

\subsection{Model Prior Setting}
Initially, the prior distributions for the observed image patches, designated as $\mathcal{H}^k$ and $\mathcal{M}^k$, are defined to follow multivariate normal distribution. By means of Eq.~\eqref{Eq:FCTN_HM}, the mean of the prior distribution for $\mathcal{H}^k$ can be expressed in terms of the factor tensors and degradation factors $\mathbf{P}_1$ and $\mathbf{P}_2$ through the $\mathbb{FCNT}$ operation. Similarly, the mean of the prior distribution for $\mathcal{M}^k$ can be described by the factor tensors and degradation factor $\mathbf{P}_3$. Moreover, the noise precision parameters for the prior distributions of $\mathcal{H}^k$ and $\mathcal{M}^k$ are represented by $\tau_h^k$ and $\tau_m^k$, respectively. Therefore, the prior distributions of $\mathcal{H}^k$ and $\mathcal{M}^k$ are given in Eq. \eqref{Eq: H_k} and Eq.~\eqref{Eq: M_k}.

\begin{figure*}
    \centering
    \begin{eqnarray}
        p\left(\mathcal{H}^{k} \big| \left \{  \mathcal{T}^{k}_{n} \right \}^4_{n=1},\tau^{k}_{h} \right) = \prod_{i'_1=1}^{I'_1} \prod_{i'_2=1}^{I'_2} \prod_{i_3=1}^{I_3} \prod_{i_4=1}^{I_4}  
        \mathcal{N}\left(h^{k}_{i'_1i'_2i_3i_4}\bigg|\mathbb{FCTN}\left ( \mathcal{T}^{k}_{1}\times_1 \mathbf{P}_1,\mathcal{T}^{k}_{2}\times_2 \mathbf{P}_2,\mathcal{T}^{k}_{3},\mathcal{T}^{k}_{4}\right)_{i'_1i'_2i_3i_4},{\tau^{k}_{h}}^{-1}\right) 
        \label{Eq: H_k}
    \end{eqnarray}
    \centering
    \begin{eqnarray}
        p\left(\mathcal{M}^{k} \big| \left \{ \mathcal{T}^{k}_{n} \right \}^4_{n=1},\tau^{k}_{m} \right) = \prod_{i_1=1}^{I_1} \prod_{i_2=1}^{I_2} \prod_{i'_3=1}^{I'_3} \prod_{i_4=1}^{I_4}  
        \mathcal{N}\left(m^{k}_{i_1i_2i'_3i_4}\bigg|\mathbb{FCTN}\left ( \mathcal{T}^{k}_{1}, \mathcal{T}^{k}_{2}, \mathcal{T}^{k}_{3} \times_3 \mathbf{P}_3, \mathcal{T}^{k}_{4} \right)_{i_1i_2i'_3i_4},{\tau^{k}_{m}}^{-1}\right)
        \label{Eq: M_k}
    \end{eqnarray}
\end{figure*}

\subsubsection{Latent Factor Tensors with Its Hyperprior}

Due to the strong correlations among factor tensors in FCTN decomposition, it is imperative that the prior settings of the factors are designed to ensure their independence and structural integrity. Hence, we assume each factor tensor $\mathcal{T}^{k}_{n}$ follows a tensor normal distribution~\cite{zeng2024bayesian} with a mean of zero and a covariance determined by a set of matrices. The prior distribution for factor tensor is represented as
\begin{equation}
    \begin{aligned}
        &p\left(\mathcal{T}^{k}_{n} \mid \boldsymbol\lambda^{k}_{1,n},\dots, \boldsymbol\lambda^{k}_{n-1,n},\boldsymbol\lambda^{k}_{n,n+1},\dots, \boldsymbol\lambda^{k}_{n,N}\right)\\
        &=\prod_{i_n}^{I_n} \mathcal{T} \mathcal{N}\left(\mathcal{T}^{k}_{n}(i_n) | \mathbf{0}, \mathbf{\Lambda}^{k}_{1,n},\dots,\mathbf{\Lambda}^{k}_{n-1,n},\mathbf{\Lambda}^{k}_{n,n+1},\dots, \mathbf{\Lambda}^{k}_{n,N}\right),
    \end{aligned}
    \label{Eq: T_n}
\end{equation}
where $\mathbf{\Lambda}^{k}_{n_1,n_2}$ = $\text{diag}(\boldsymbol{\lambda}^{k}_{n_1,n_2})$, $\text{diag}(\cdot )$ denotes the matrixization of a vector and $\mathcal{T} \mathcal{N}(\cdot )$ denotes tensor normal distribution. The $\boldsymbol{\lambda}^{k}_{n_1,n_2}$ is a parameter to establish a connection between the factor tensors $ \mathcal{T}^{k}_{n_1} $ and $ \mathcal{T}^{k}_{n_2} $. 

The structure of the covariance of the above distribution is complex and non-intuitive.  Fortunately, with the rule of linear transformation from tensor normal distribution to multivariate normal distribution\footnote{$\mathcal{X} \sim \mathcal{TN}(\mathcal{M}, \Sigma_1, \Sigma_2, \ldots, \Sigma_k) \Leftrightarrow  \text{vec}(\mathcal{X}) \sim \mathcal{N}(\text{vec}(\mathcal{M}), \Sigma_k \otimes \Sigma_{k-1} \otimes \cdots \otimes \Sigma_1)$\cite{kolda2009tensor}. }, this covariance can be replaced by the matricization for the Kronecker product of a set of vectors. Thus, the prior distribution can be expressed as a more intuitive multivariate normal distribution as follows:
\begin{equation}
    \begin{aligned}
        &p\left(\mathcal{T}^{k}_{n} \mid \boldsymbol\lambda^{k}_{1,n},\dots, \boldsymbol\lambda^{k}_{n-1,n},\boldsymbol\lambda^{k}_{n,n+1},\dots, \boldsymbol\lambda^{k}_{n,N}\right)\\
        & \quad=\prod_{i_n}^{I_n} \mathcal{N}\left({\mathbf{T}^{k}_{n}}_{(n)}(i_n) \mid \mathbf{0}, \mathbf{\Lambda}^{k}_{n}\right),
    \end{aligned}
    \label{Eq: T_n_unfold}
\end{equation}
where $\boldsymbol{\Lambda}^{k}_{n} = \text{diag}(\boldsymbol{\lambda}^{k}_{n,N}\odot \cdots \odot \boldsymbol{\lambda}^{k}_{n,n+1}\odot \boldsymbol{\lambda}^{k}_{n,n-1}\odot \cdots \odot 
\boldsymbol{\lambda}^{k}_{1,n})$.

With the above, each pair of factor tensors $\mathcal{T}^{k}_{n_1}$ and $\mathcal{T}^{k}_{n_2}$  are interconnected by $\boldsymbol{\lambda}^{k}_{n_1,n_2} $, adhering strictly to the FCTN decomposition paradigm and forming a fully connected tensor network. The parameter $\boldsymbol{\lambda}^{k}_{n_1,n_2} $ for establishing the connection of factors is considered to follow a Gamma distribution, as

\begin{equation}
    \begin{aligned}
        p\left(\boldsymbol{\lambda}^{k}_{n_1,n_2}\right) =\prod_{r}^{R_{n_1,n_2}} \text{Ga}\left((\lambda^{k}_{n_1,n_2})_r\mid a_0,b_0\right),
    \end{aligned}
    \label{Eq:lambda_n1_n2}
\end{equation}
where $\text{Ga}(x\mid a,b) = \frac{b^{a}x^{a-1}e^{-bx}}{\Gamma{(a)}}$ represents the expression for the Gamma distribution and $\mathbb{E}(x) = \frac{a}{b}$, the $a_0$ and $b_0$ represent shape and rate parameters, respectively. The Gamma prior is defined on the positive real domain, aligning with the role of $\boldsymbol{\lambda}^{k}_{n_1,n_2}$ as a non-negative association control parameter between factor tensors. Moreover, it allows conjugate updates within the exponential family, which facilitates efficient inference~\cite{kruschke2010bayesian}.

\subsubsection{Precision Parameters}
\begin{figure} [h!]
    \includegraphics[width=\linewidth]{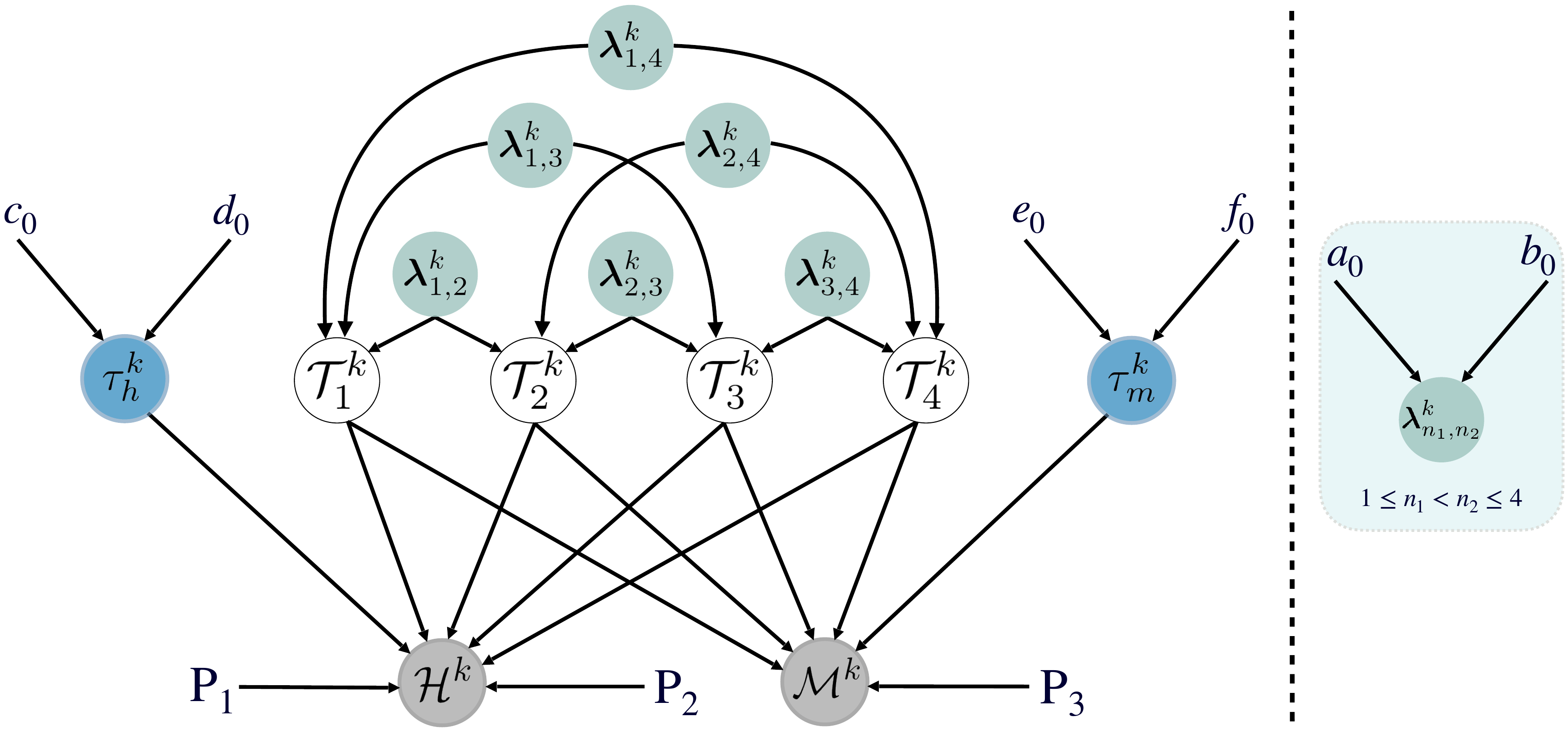}
    \caption{The probabilistic graphical model of the proposed BFCTN model. The left part is the overall framework of the probabilistic graphical model, and the right part is the details of the hyperparameters of the factor tensors.}
    \label{fig:Model}
\end{figure}

To ensure the proposed model within a fully Bayesian framework and improve its robustness, the precision parameters ${\tau^{k}_{h}}$ and ${\tau^{k}_{m}}$ in Eq.~\eqref{Eq: H_k} and Eq.~\eqref{Eq: M_k} are also assumed to follow a Gamma distribution, as
\begin{equation}
    \begin{aligned}
        p({\tau^{k}_{h}}) = \text{Ga}\left({\tau^{k}_{h}}\mid c_0,d_0\right),
    \end{aligned}
    \label{Eq:alpha}
\end{equation}
\begin{equation}
    \begin{aligned}
        p({\tau^{k}_{m}}) = \text{Ga}\left({\tau^{k}_{m}}\mid e_0,f_0\right).
    \end{aligned}
    \label{Eq:beta}
\end{equation}
Here, $c_0$ and $e_0$ are the shape parameters for these distributions, and $d_0$ and $f_0$ are the rate parameters. At this point, a fully Bayesian FCTN model for hyperspectral-multispectral image fusion has been formulated, with the probabilistic graph model presented in Fig.~\ref{fig:Model}.

\subsubsection{Joint Distribution}
To simplify the representation, we use a set to represent all latent variables and hyperparameters in our proposed model, which is designated $\boldsymbol{\Xi} = \left \{ \left \{  \mathcal{T}^{k}_{n} \right \}^4_{n=1}, \left \{\boldsymbol{\lambda}^{k}_{n_1,n_2}\right \}_{1\le n_1<n_2\le4}, \tau^{k}_{h}, \tau^{k}_{m}, \right \}$.  
As illustrated in the probabilistic graphical models in Fig.~\ref{fig:Model}, the joint distribution of the BFCTN model can be written as
\begin{equation}
    \begin{aligned}
        & p\left(\boldsymbol{\Xi}^{k}\right)=p\left( \mathcal{H}^{k} \big| \left \{  \mathcal{T}^{k}_{n} \right \}^4_{n=1},\tau_h^k \right)p\left(\mathcal{M}^{k} \big| \left \{ \mathcal{T}^{k}_{n} \right \}^4_{n=1},\tau_m^k \right)\\
        &\qquad\quad \prod_{n=1}^{4} p\left(\mathcal{T}^{k}_{n} \mid \boldsymbol\lambda^{k}_{1,n},\dots, \boldsymbol\lambda^{k}_{n-1,n},\boldsymbol\lambda^{k}_{n,n+1},\dots, \boldsymbol\lambda^{k}_{n,N}\right) \\
        &\qquad\quad \prod_{n_1=1}^{3} \prod_{n_2=2}^{4} p\left(\boldsymbol\lambda^{k}_{n_1, n_2}\right)p\left(\tau^{k}_{h}\right)p\left(\tau^{k}_{m}\right).
    \end{aligned}
    \label{Eq:joint}
\end{equation}

By combining the likelihood terms in Eq.~\eqref{Eq: H_k} and Eq.~\eqref{Eq: M_k}, the priors of factor tensors and their covariance (given in Eq.~\eqref{Eq: T_n_unfold} and Eq.~\eqref{Eq:lambda_n1_n2}), and the hyperpriors of the noise precision parameter (described in Eq.~\eqref{Eq:alpha} and Eq.~\eqref{Eq:beta}), the log form of the joint distribution can be expressed as follows:

\begin{equation}
    \begin{aligned}
        & \ell(\boldsymbol{\Xi}^{k})=  -\frac{\tau^{k}_{h}}{2}\left\|\mathcal {H}^{k}-\mathbb{FCTN}\left ( \mathcal{T}^{k}_1\times_1 \mathbf{P}_1,\mathcal{T}^{k}_2\times_2 \mathbf{P}_2,\mathcal{T}^{k}_3 ,\mathcal{T}^{k}_4\right)\right\|_{\mathrm{F}}^2\\ 
        & \quad\quad\quad -\frac{\tau^{k}_{m}}{2} \left\|\mathcal{M}^{k}-\mathbb{FCTN}\left (\mathcal{T}^{k}_1,\mathcal{T}^{k}_2,\mathcal{T}^{k}_3 \times_3 \mathbf{P}_3,\mathcal{T}^{k}_4\right) \right\|_{\mathrm{F}}^2  \\
        & \quad\quad\quad +\sum_{n=1}^{4}\sum_{i_n}^{I_n} \left( \frac{1}{2}\ln\left | \boldsymbol{\Lambda}^{k}_{n} \right |  -\frac{1}{2}{\mathbf{T}^{k}_{n}(i_n) }^{\top}\mathbf{\Lambda}^{k}_{n} {\mathbf{T}^{k}_{n}(i_n) }\right) \\
        & \quad\quad\quad +\sum_{n_1=1}^{3}\sum_{n_2=2}^{4}\sum_{r}^{R_{n_1,n_2}}\left(a_0-1\right) \ln (\lambda^{k}_{n_1,n_2})_r\\
        & \quad\quad\quad -\sum_{n_1=1}^{3}\sum_{n_2=2}^{4}\sum_{r}^{R_{n_1,n_2}}b_0 (\lambda^{k}_{n_1,n_2})_r\\
        & \quad\quad\quad +\left(\frac{I'_1I'_2I_3I_4}{2}+c_0-1\right) \ln \tau^{k}_{h}\\
        & \quad\quad\quad +\left(\frac{I_1I_2I'_3I_4}{2} + e_0-1\right) \ln \tau^{k}_{m}\\
        & \quad\quad\quad -d_0 \tau^{k}_{h} - f_0 \tau^{k}_{m} + C.
    \end{aligned}
    \label{Eq:ln_joint}
\end{equation}

Here, $C$ is a constant.

\subsection{Model Inference}
This section presents a comprehensive account of the model inference process. The principal objective of our model is to estimate the latent factor tensors $\left\{\mathcal{T}^{k}_{n}\right\}^{4}_{n=1}$ to reconstruct the fused HR-HSI patches. To achieve this objective, the estimation of the parameters $\{\boldsymbol\lambda^{k}_{n_1, n_2}\}_{1\le n_1<n_2\le4}$ and noise precision parameters $\tau^{k}_{h}$ and $\tau^{k}_{m}$ are equally crucial.

Exact Bayesian inference, despite providing theoretically optimal probability estimates, is often computationally unfeasible in practice, especially for high-order observed data and complex models. Fortunately, variational Bayesian approaches provide an efficient approximate solution \cite{zhao2015bayesian, zhang2018advances, zhao2015robust}. In variational inference, a more tractable distribution is employed to approximate the complex and intractable posterior distribution, with the Kullback-Leibler (KL) divergence utilized to quantify the divergence between the two distributions.

Therefore, we seek to approximate the true posterior distribution $p(\boldsymbol{\Xi}^{k} \mid \mathcal{H}^{k},\mathcal{M}^{k})$ in our model by finding a tractable distribution $q(\boldsymbol{\Xi}^{k})$ through the minimization of the KL divergence, that is

\begin{equation}
    \begin{aligned}
        \mathrm{KL}(q&(\boldsymbol{\Xi}^{k}) \| p(\boldsymbol{\Xi}^{k}\mid\mathcal{H}^{k},\mathcal{M}^{k})) =  \ln(p(\mathcal{H}^{k},\mathcal{M}^{k})) \\
        &- \left(\mathbb{E}_{q(\boldsymbol{\Xi}^{k})} \left[\ln p(\boldsymbol{\Xi}^{k},\mathcal{H}^{k},\mathcal{M}^{k})\right] - \mathbb{E}_{q(\boldsymbol{\Xi}^{k})} \left[\ln q(\boldsymbol{\Xi}^{k})\right] \right), \\
    \end{aligned}
    \label{Eq:KL}
\end{equation}
where the term $\ln(p(\mathcal{H}^{k},\mathcal{M}^{k}))$ is a constant, representing the model evidence, $\mathbb{E}_{q(\boldsymbol{\Xi}^{k})} [\ln p(\boldsymbol{\Xi}^{k},\mathcal{H}^{k},\mathcal{M}^{k})] - \mathbb{E}_{q(\boldsymbol{\Xi}^{k})} [\ln q(\boldsymbol{\Xi}^{k})]$ is called evidence lower bound ($ELBO$), and due to the non-negative nature of KL divergence, KL divergence will approach 0 when $ELBO$ reaches the bound of evidence.

Based on the mean-field approximation assumption, all latent variables in $\boldsymbol{\Xi}^{k}$ can be considered independent of each other, allowing the variational distribution to be factorized as
\begin{equation}
    \begin{aligned}             
        q\left(\boldsymbol{\Xi}^{k}\right)=\prod_{n=1}^{4} q\left(\mathcal{T}^{k}_{n} \right) \prod_{n_1=1}^{3} \prod_{n_2=2}^{4} q\left(\boldsymbol{\lambda}^{k}_{n_1, n_2}\right)  q\left(\tau^{k}_{h}\right)q\left(\tau^{k}_{m}\right).
    \end{aligned}
\end{equation}

With this, the optimal form for each latent variable can be obtained from 
\begin{equation}
    \begin{aligned}
        \ln q_{i}\left(\boldsymbol{\Xi}^{k}_{i}\right)=\mathbb{E}_{q_{(\boldsymbol{\Xi}^{k}\setminus \boldsymbol{\Xi}^{k}_{i})}}\left [ \ln p(\mathcal{H}^{k},\mathcal{M}^{k},\boldsymbol{\Xi}^{k}  ) \right ]+{C},
        \label{Eq:lnq_i}
    \end{aligned}
\end{equation}
where $q_{i}\left(\boldsymbol{\Xi}^{k}_{i}\right)$ denotes the estimated posterior distribution of the 
$i$-th variable in $\boldsymbol{\Xi}$, and $\mathbb{E}_{q_{(\boldsymbol{\Xi}^{k}\setminus \boldsymbol{\Xi}^{k}_{i})}}$ denotes the expectation over the distribution of all variables except for the $i$-th variable. Since the prior distributions of all latent variables in the above are drawn from the exponential family, we can deriving the posterior distribution for each variable from Eq.~\eqref{Eq:lnq_i}.

\subsubsection{Estimation of Latent Factor Tensors}
As shown in Fig.~\ref{fig:Model}, the latent factor tensor $\mathcal{T}^{k}_n$ receives the messages from the observed image patches $\mathcal{H}^{k}$ and $\mathcal{M}^{k}$, and its co-parents, including the covariance of all factor tensors. With the Eq.~\eqref{Eq:lnq_i}, the posterior distribution of $\mathcal{T}^{k}_n$ can be constructed by the likelihood terms in Eq.~\eqref{Eq: H_k}-\eqref{Eq: M_k}, and the prior in Eq.~\eqref{Eq: T_n_unfold}.

However, the difference in dimensions between image patches $\mathcal{H}^{k}$ and $\mathcal{M}^{k}$ presents a challenge in factoring the latent factor tensors into independent variables. Fortunately, with appropriate regularization~\cite{cheng2020learning}, the $\mathcal{T}^{k}_n$ can be constrained by applying a Dirac delta function, resulting in $Q\left(\mathcal{T}^{k}_n\right)=\delta\left(\mathcal{T}^{k}_n-\hat{\mathcal{T}}^{k}_n\right)$ as the point estimate. This allows the variational problem of minimizing the KL divergence to be transformed into an expectation-maximization problem, as follows
\begin{equation}
    \begin{aligned}
        \hat{\mathcal{T}^{k*}_n}=\arg \max \mathbb{E}_{\Pi_{\boldsymbol{\Xi}_{i} \neq \mathcal{T}^{k}_n} Q\left(\boldsymbol{\Xi}_{i}\right)}[\ln p(\boldsymbol{\Xi}^{k})].
    \end{aligned}
    \label{Eq:T*}
\end{equation}

The EM algorithm can be employed to treat the latent factor tensor as a global variable~\cite{meng1993maximum, ye2023bayesian}, with the remaining variables designated as latent parameters. This allows for the alternation of updates to each latent factor tensor.
Taking $\mathcal{T}^{k}_1$ as an example, its point estimate can be viewed as the result of optimizing the following convex function:
\begin{equation}
    \begin{aligned}
        & \max \mathbb{E}_{q\left(\boldsymbol{\Xi}^{k} \backslash \mathcal{T}^{k}_1\right)}\\
        &\quad \Bigg\{ -\frac{\tau_h^{k}}{2}\left\|\mathcal {H}^{k}-\mathbb{FCTN}\left ( \mathcal{T}^{k}_1\times_1 \mathbf{P}_1,\mathcal{T}^{k}_2\times_2 \mathbf{P}_2,\mathcal{T}^{k}_3 ,\mathcal{T}^{k}_4\right)\right\|_{\mathrm{F}}^2\\
        & \quad -\frac{\tau_m^{k}}{2}\left\|\mathcal{M}^{k}-\mathbb{FCTN}\left ( \mathcal{T}^{k}_1,\mathcal{T}^{k}_2,\mathcal{T}^{k}_3 \times_3 \mathbf{P}_3,\mathcal{T}^{k}_4\right) \right\|_{\mathrm{F}}^2\\
        & \quad-\frac{1}{2} \operatorname{Tr}\left(\mathbf{T}^{k}_{1(1)} \mathbf{\Lambda}^{k}_{1}{\mathbf{T}^{k}_{1(1)}}^{\top}\right)\Bigg\}. \\
    \end{aligned}
    \label{Eq:maxE_T}
\end{equation}
By optimizing the convex function in Eq.~\eqref{Eq:maxE_T}, we can obtain a Sylvester equation\footnote{ $\mathbf{A} \mathbf{X} \mathbf{B}+\mathbf{C} \mathbf{X} \mathbf{D}=\mathbf{E} \Rightarrow  \boldsymbol{\mathbf{x}}= \left(\mathbf{B} \otimes \mathbf{A}+\mathbf{D} \otimes \mathbf{C}\right)^{-1}\boldsymbol{\mathbf{e}}$, where $\mathbf{x}$ and $\mathbf{e}$ represent vectorization of $\mathbf{X}$ and $\mathbf{E}$, respectively~\cite{ye2023bayesian, simoncini2016computational}.} as follows:

\begin{equation}
    \begin{aligned}
        &\mathbf{T}^{k}_{1(1)}\left\{\mathbb{E}\left[\tau^{k}_{m}\right] \mathbb{E}\left[{\mathbf{M}^{k}_{\neq1}}^\top\mathbf{M}^{k}_{\neq1}\right]+\mathbb{E}\left[\boldsymbol{\Lambda}^{k}_1\right]\right\} \\
        &+\mathbb{E}\left[\tau^{k}_{h}\right] \mathbf{P_1}^{\top} \mathbf{P_1} \mathbf{T}^{k}_{1(1)} \mathbb{E}\left[{\mathbf{H}^{k}_{\neq1}}^{\top}\mathbf{H}^{k}_{\neq1}\right] \\
        &=\mathbb{E}\left[\tau_{m}^{k}\right]\mathbf{M}^{k}_{(1)} \mathbb{E}\left[{\mathbf{M}^{k}_{\neq1}}^{\top}\right]+\mathbb{E}\left[\tau_{h}^{k}\right] \mathbf{P_1}^{\top}\mathbf{H}^{k}_{(1)} \mathbb{E}\left[{\mathbf{H}^{k}_{\neq1}}^{\top}\right] . \\
    \end{aligned}
    \label{Eq:Sylvester}
\end{equation}
The point estimates of the latent factor tensor $\mathcal{T}^{k}_{1}$ can be obtained by solving the above Sylvester equation. The most intricate terms in this equation are $\mathbf{H}^{k}_{\neq1}$ and $\mathbf{M}^{k}_{\neq1}$, and we explain them as follows.
With $\hat{\mathcal{H}}^{k} = \mathbb{FCTN}\left ( \mathcal{T}^{k}_1\times_1 \mathbf{P}_1,\mathcal{T}^{k}_2\times_2 \mathbf{P}_2,\mathcal{T}^{k}_3 ,\mathcal{T}^{k}_4\right)$ and using Eq.~\eqref{Eq:X_n},  we obtain the intricate term $\mathbf{H}^{k}_{\neq1}$ in the following equation $\hat{\mathbf{H}}^{k}_{(1)}= \mathbf{T}^{k}_{1(1)}\mathbf{H}^{k}_{\neq1}$.
Similarly, let $\hat{\mathcal{M}}^{k} = \mathbb{FCTN}\left ( \mathcal{T}^{k}_1,\mathcal{T}^{k}_2,\mathcal{T}^{k}_3 \times_3 \mathbf{P}_3,\mathcal{T}^{k}_4\right)$ and using Eq.~\eqref{Eq:X_n}, we obtain the $\mathbf{M}^{k}_{\neq1}$ in $\hat{\mathbf{M}}^{k}_{(1)}= \mathbf{T}^{k}_{1(1)}\mathbf{M}^{k}_{\neq1}$.
With the above approach, the factor tensor $\mathcal{T}^{k}_1$ can be factored into independent variables, yielding its point estimate. The other factor tensors are estimated in a similar manner by constructing and solving their corresponding Sylvester equations detailed in the Appendix. \ref{appendix: factor}.

\subsubsection{Estimation of Hyperprior}
As shown in Fig.~\ref{fig:Model}, the parameters $\{\boldsymbol{\lambda}_{n_1, n_2}^{k}\}$ establishes the connection between each factor tensor to form a full connected tensor network. The $\boldsymbol{\lambda}_{n_1, n_2}^{k}$ receives messages from its hyperprior and associated factor tensors, including $\mathcal{T}_{n_1}^{k}$, and $\mathcal{T}_{n_2}^{k}$. By applying Eq.~\eqref{Eq:lnq_i}, it has been shown that the posterior distribution of $\boldsymbol{\lambda}_{n_1, n_2}^{k}$ can be factorized as independent Gamma distributions (see Appendix. \ref{appendix: lambda} for more details), as
\begin{equation}
    q\left(\boldsymbol{\lambda}^{k}_{n_1, n_2}\right)=\prod_{r}^{R_{n_1, n_2}} \text{Ga}\left((\lambda^{k}_{n_1, n_2})_r\mid a^{k}_{n_1,n_2}, b^{k}_{n_1,n_2}\right),
    \label{Eq:q_lambda}
\end{equation}
with posterior parameters updated by
\begin{equation}
    \begin{aligned}
        & a^{k}_{n_1,n_2}=a_0+\frac{I_{n_1}\prod_{\substack{n\neq{n_1,n_2}}}^{N}R_{n,n_1} + I_{n_2}\prod_{\substack{n\neq{n_1,n_2}}}^{N}R_{n,n_2} }{2}\\
        & b^{k}_{n_1,n_2}=b_0 +  \frac{1}{2}\left (  \mathbb{E}_{q}\left[{\mathbf{T}^{k}_{n_1(n_2)}(r) }^{\top}\boldsymbol{\lambda}^{k}_{n_1,\neq n_2} {\mathbf{T}^{k}_{n_1(n_2)}(r) } \right]\right.\\
        &\quad \left.\quad + \mathbb{E}_{q}\left[{\mathbf{T}^{k}_{n_2(n_1)}(r)}^{\top}\boldsymbol{\lambda}^{k}_{n_2,\neq n_1} {\mathbf{T}^{k}_{n_2(n_1)}(r) } \right]\right ),
    \end{aligned}
    \label{Eq:a_k,b_k}
\end{equation}
where $\boldsymbol{\lambda}^{k}_{n_1,\neq n_2} = \boldsymbol{\lambda}^{k}_{n_1,N}\odot \cdots \odot \boldsymbol{\lambda}^{k}_{n_1,n_2+1}\odot \boldsymbol{\lambda}^{k}_{n_1,n_2-1}\odot \cdots \odot \boldsymbol{\lambda}^{k}_{n_1,n_1+1}\odot \boldsymbol{\lambda}^{k}_{n_1,n_1-1}\odot\cdots \odot  \boldsymbol{\lambda}^{k}_{1,n_1}$. Additionally, $\mathbf{T}^{k}_{n_1(n_2)}$ represents the matrix form of the factor tensor $\mathcal{T}_{n_1}^{k}$ unfolded along the $n_2$-th model. According to the FCTN decomposition defined in Eq. \eqref{FCTN}, the size of first dimension of $\mathbf{T}^{k}_{n_1(n_2)}$ is $R_{n_1,n_2}$.
The parameter $a^{k}_{n_1,n_2}$ depends on the dimensions and tensor ranks associated with the subscripts $n_1$ and $n_2$, while $b^{k}_{n_1,n_2}$ is strongly correlated with the values of $\mathcal{T}^{k}_{n_1}$ and $\mathcal{T}^{k}_{n_2}$.

\subsubsection{Estimation of Noise Precision Parameters}
The inference of the noise precision $\tau_h^k$ is performed by receiving messages from its hyperprior and the observed image patches $\mathcal{H}^{k}$ and hyperparameter $c_0$. 
Applying Eq. \eqref{Eq:lnq_i}, the posterior distribution of $\tau^{k}_{h}$ follows a Gamma distribution (see Appendix. \ref{appendix: tau_h} for more details), given by
\begin{equation}
    \begin{aligned}
        q(\tau^{k}_{h})=\text{Ga}\left(\tau^{k}_{h} \mid c^k, d^k\right),
    \end{aligned}
    \label{Eq:q_alpha}
\end{equation}
where the posterior parameters can be updated by
\begin{equation}
    \begin{aligned}
        & c^{k}=c_0+\frac{I'_1 I'_2 I_3 I_4}{2} \\
        & d^{k}=d_0+ \\
        &\: \frac{1}{2} \mathbb{E}_q\left[\left\|\left({\mathcal {H}^{k}}-\mathbb{FCTN}\left(\mathcal{T}^{k}_1 \times_1 \mathbf{P}_1, \mathcal {T}^{k}_2 \times_2 \mathbf{P}_2, \mathcal{T}^{k}_3, \mathcal{T}^{k}_4\right)\right)\right\|_F^2\right].
    \end{aligned}
    \label{Eq:c_k,d_k}
\end{equation}
In the above, the posterior parameter of $c^{k}$  is determined by the dimensions of $\mathcal{H}^{k}$ and hyperparameter $c_0$. The term $\left\|\left({\mathcal {H}^{k}}-\mathbb{FCTN}\left(\mathcal{T}^{k}_1 \times_1 \mathbf{P}_1, \mathcal {T}^{k}_2 \times_2 \mathbf{P}_2, \mathcal{T}^{k}_3, \mathcal{T}^{k}_4\right)\right)\right\|_F^2$ can be interpreted as the residual between the after-downsampling in the spatial dimension of the fused image patches and the observed $\mathcal{H}^{k}$.

Similarly, the posterior distribution of the noise precision $\tau^{k}_{m}$ also follows a Gamma  distribution (see Appendix. \ref{appendix: tau_m} for more details), as
\begin{equation}
    \begin{aligned}
        q(\tau^{k}_{m})=\text{Ga}\left(\tau^{k}_{m} \mid e^k, f^k\right),
    \end{aligned}
    \label{Eq:q_beta}
\end{equation}
with posterior parameters updated by
\begin{equation}
    \begin{aligned}
        & e^{k}=e_0+\frac{I_1 I_2 I'_3 I_4}{2} \\
        & f^{k}=f_0+ \\
        & \quad\frac{1}{2} \mathbb{E}_q\left[\left\|\left({\mathcal { M }^{k} }-\mathbb{FCTN}\left(\mathcal{T}^{k}_1, \mathcal {T}^{k}_2, \mathcal{T}^{k}_3 \times_3 \mathbf{P}_3, \mathcal{T}^{k}_4\right)\right)\right\|_F^2\right].
    \end{aligned}
    \label{Eq:e_k,f_k}
\end{equation}

From Eq. \eqref{Eq:e_k,f_k}, it is clear that the parameter $e^k$ is derived from the dimensional of the observed $\mathcal{M}^k$ and remains constant, while $f^k$ represents the residual between the fused and observed image patches.
According to the properties of the Gamma distribution, smaller residuals in Eq.~\eqref{Eq:c_k,d_k} and \eqref{Eq:e_k,f_k} lead to higher expectation values of the corresponding precision parameters, which also reflects better fusion performance.

\subsubsection{Lower Bound of Model Evidence}
The proposed variational inference framework maximizes the $ELBO$ defined in Eq.~\eqref{Eq:KL}, which serves as the theoretical objective for convergence. The $ELBO$ expression in Eq.~\eqref{Eq:ELBO} consists of two components:
\begin{equation}
    ELBO=\mathbb{E}_{q(\Xi^k )}\left[\ln p\left(\mathcal{H}^k,\mathcal{M}^k, \Xi^k \right)\right]+H(q(\Xi^k )),
    \label{Eq:ELBO}
\end{equation}
where the first term is the expected joint log-likelihood and the second term represents the entropy of the variational distribution. Under the exponential family assumptions and conjugate priors, a closed-form expression of the $ELBO$ is derived in Eq.~\eqref{Eq:ELBO2}, where all terms are analytically tractable:

\begin{equation}
    \begin{aligned}
     &ELBO \\
     & = -\frac{c^k}{2 d^k} \mathbb{E}_{q}\left[\left\|\mathcal {H}^{k}-\mathbb{FCTN}\left ( \mathcal{T}^{k}_1\times_1 \mathbf{P}_1,\mathcal{T}^{k}_2\times_2 \mathbf{P}_2,\mathcal{T}^{k}_3 ,\mathcal{T}^{k}_4\right)\right\|_{F}^{2}\right] \\
     &\quad -\frac{e^k}{2 f^k} \mathbb{E}_{q}\left[\left\|\mathcal {M}^{k}-\mathbb{FCTN}\left (\mathcal{T}^{k}_1,\mathcal{T}^{k}_2,\mathcal{T}^{k}_3 \times_3 \mathbf{P}_3,\mathcal{T}^{k}_4\right) \right\|_{\mathrm{F}}^2\right] \\
     &\quad-\frac{1}{2}\sum_{n=1}^{4}\mathrm{Tr}\left\{\mathbb{E}_q[\mathbf{\Lambda}^k_n]\left({\mathbf{T}^k_{n(n)}}^{\top}\mathbf{T}^k_{n(n)}\right)\right\} \\
     &\quad+\sum_{n_1=1}^{3}\sum_{n_2=n_1+1}^{4} \sum_{r}^{R_{n_1,n_2}}\Bigg\{\ln\Gamma(a_{n_1,n_2}^k)\\
     &\quad+a_{n_1,n_2}^k\left(1-\ln b_{n_1,n_2}^k-\frac{b_0^k}{b_{n_1,n_2}^k}\right)\Bigg\}\\
     &\quad+\ln\Gamma(c^k)+c^k(1-\ln d^k-\frac{d^k_0}{d^k})\\
     &\quad+\ln\Gamma(e^k)+e^k(1-\ln f^k-\frac{f^k_0}{f^k}) + C.
    \end{aligned}
    \label{Eq:ELBO2}
\end{equation}	

The structure of Eq.~\eqref{Eq:ELBO2} offers an intuitive interpretation. The first two terms characterize the expected reconstruction errors of $\mathcal{H}^k$ and $\mathcal{M}^k$, reflecting the model's fidelity to the observations. The third term imposes regularization through the factor tensor priors. The remaining terms quantify the divergence between posteriors and hyperpriors, serving as a natural penalty to control model complexity. In practice, the balance among these terms reflects the trade-off between data fidelity and model regularity during optimization (see Appendix. \ref{appendix: ELBO} for more details).

\subsubsection{Model Implementation and Complexity Analysis}
The implementation of the proposed BFCTN model for HMF proceeds as follows. Firstly, the observed LR-HSI $\mathcal{H}$ and HR-MSI $\mathcal{M}$ are chunked into overlapping patches of matched sizes and divided into $K$ groups based on their column-wise spatial locations, yielding $\mathcal{H}^k$ and $\mathcal{M}^k$ for each group. Secondly, for each group $k$, the model introduces four latent factor tensors $\{\mathcal{T}^{k}_n\}_{n=1}^{4}$ to capture shared spectral and spatial structures. These latent factors are inferred through a Bayesian process involving the estimation of parameters $\{\boldsymbol{\lambda}^k_{n_1, n_2}\}_{1\le n_1<n_2\le4}$ governing their distributions, as well as noise precision parameters $\tau_h^k$ and $\tau_m^k$ for modeling the observation noise in $\mathcal{H}^k$ and $\mathcal{M}^k$, respectively. The fused HR-HSI patches $\{\mathcal{Z}^k\}_{k=1}^K$ are then obtained via the FCTN combination mechanism, inheriting enhanced spatial details from $\mathcal{M}^k$ and preserving spectral fidelity from $\mathcal{H}^k$. Finally, the full-resolution HR-HSI $\mathcal{Z}$ is reconstructed by aggregating $\{\mathcal{Z}^k\}$ through averaging overlapping regions to ensure spatial consistency and smooth transitions. The proposed model is summarized in Algorithm~\ref{alg:BFCTN}.

To assess computational feasibility, the time complexity is analyzed as follows. Assuming all factor ranks are equal, i.e., $R_{n_1,n_2} = R$ for all $1 \leq n_1 < n_2 \leq N$, and all mode dimensions satisfy $I_1 = I_2 = \dots = I_N = I$, the computational cost mainly arises from the combination and estimation of factor tensors. The combination in Eqs.~\eqref{Eq:q_alpha} and \eqref{Eq:q_beta} incurs $\mathcal{O}\left(N \sum_{n=2}^{N-1} R^{n(N-n)+n-1} I^n + N R^{2(N-1)} I + N R^{3(N-1)}\right)$ complexity. The estimation via solving the Sylvester equation in Eq.~\eqref{Eq:Sylvester} contributes $\mathcal{O}\left(N R^{2(N-1)} + N R^{3(N-1)} + N I^3\right)$, assuming standard matrix equation solvers. In the BFCTN framework with fixed tensor order $N = 4$, the overall complexity for processing $K$ groups simplifies to $\mathcal{O}\left(K I^3 R^6 + K R^9\right)$. This reflects the trade-off between reconstruction fidelity and computational cost inherent in BFCTN.

\begin{algorithm}
    \caption{BFCTN for LR-HSI and HR-MSI Fusion}
    \label{alg:BFCTN}
    \begin{algorithmic}[1]
        \Require $\mathcal{H}, \mathcal{M}, \mathbf{P}_1, \mathbf{P}_2, \mathbf{P}_3, sf$.
        \Ensure High-resolution hyperspectral image $\mathcal{Z}$.
        \State Chunk $\mathcal{H}$ and $\mathcal{M}$ and divide into $K$ groups.
        \State Initialize hyperparameters $a_0, b_0, c_0, d_0, e_0, f_0$, and parameters $\{\mathcal{T}^{k}_n\}_{n=1}^{4}, \{\boldsymbol\lambda^{k}_{n_1,n_2}\}, \tau^{k}_{h}, \tau^{k}_{m}$.
        \For{$k = 1,..., K$}
        \While{\textit{Iterations < max-iterations}}
        \State/* \textit{ 1) Estimate covariance paermeters} */
        \For{$n_1=1,..., 3$}
        \For{$n_2$=$n_1+1,..., 4$}
        \State Compute $\hat{a}^{k}_{n_1,n_2}$ and $\hat{b}^{k}_{n_1,n_2}$ via Eq. \eqref{Eq:a_k,b_k};
        \State Update $q(\boldsymbol\lambda^{k}_{n_1,n_2})$ by Eq. \eqref{Eq:q_lambda};
        \EndFor
        \EndFor
        \State/* \textit{ 2) Estimate noise precision parameters} */
        \State Compute $\hat{c}^{k}$ and $\hat{d}^{k}$ via Eq. \eqref{Eq:c_k,d_k}; 
        \State Update the posterior $q(\tau^{k}_{h})$ by Eq. \eqref{Eq:q_alpha}; 
        \State Compute $\hat{e}^{k}$ and $\hat{f}^{k}$ via Eq. \eqref{Eq:e_k,f_k};
        \State Update the posterior $q(\tau^{k}_{m})$ by Eq. \eqref{Eq:q_beta};
        \State/* \textit{ 3) Estimate latent factor tensors} */
        \For{$n=1,...,4$}
        \State Establish the Sylvester equation as Eq. \eqref{Eq:Sylvester};
        \State Update the point estimation of $\mathcal{T}^{k}_n$;
        \EndFor 
        \EndWhile
        \State Construct $\mathcal{Z}^k$ by $\mathbb{FCTN}\left(\mathcal{T}^{k}_1, \mathcal {T}^{k}_2, \mathcal{T}^{k}_3, \mathcal{T}^{k}_4\right)$.
        \EndFor
        \State Combine $\{\mathcal{Z}^k\}_{k=1}^{K}$ into $\mathcal{Z}$.
    \end{algorithmic}
\end{algorithm}

\section{Experiments and Discussion}
\label{sec:exp}
This section presents the results of extensive experiments and a comprehensive analysis conducted to evaluate the performance of the proposed method. Experiments on conventional methods are run using MATLAB 2023a on a PC equipped with a 1.4 GHz quad-core Intel Core i5 processor and 16 GB of 2133 MHz LPDDR3 memory, while experiments involving deep learning based methods are conducted on a system with an M3 MAX processor and 64 GB of unified RAM.

\begin{table*}[t!]
    \caption{The fusion performance of different methods in different scenarios is compared on the CAVE dataset.}
    \begin{tabular}{ccccccccccccc}
        \hline
        \multirow{2}{*}{SNR=35dB} & \multicolumn{4}{c}{sf=4}                                               & \multicolumn{4}{c}{sf=8}                                               & \multicolumn{4}{c}{sf=16}                                              
        \\ \cmidrule(lr){2-5}  \cmidrule(lr){6-9}  \cmidrule(lr){10-13} 
        & PSNR$\uparrow$  & SSIM$\uparrow$ &ERGAS$\downarrow$ &SAM $\downarrow$ & PSNR $\uparrow$  & SSIM$\uparrow$  & ERGAS$\downarrow$ & SAM$\downarrow$ & PSNR$\uparrow$  & SSIM $\uparrow$ & ERGAS$\downarrow$ & SAM$\downarrow$    \\ \hline
        LTTR              & 40.5335         & 0.9732         & 2.3478           & 15.3737          & 40.4722          & 0.9705          & 1.2882          & 16.1115           & 39.8726         & 0.9692          & 0.7377          & 16.3582          \\
        LTMR              & 40.4721         & 0.9735         & 2.3373           & 14.4449          & 40.3315          & 0.9705          & 1.3025          & 15.0552           & 39.7673         & 0.9671          & 0.7360          & 15.6794           \\
        GTNN              & 40.1069         & 0.9693         & 2.4413           & 15.5963            & 40.2650          & 0.9714          & 1.3187          & 14.7387         & 39.7292         & 0.969           & 0.7475          & 14.8752           \\
        CNMF              & 43.6854         & 0.9790         & 1.7442           & 14.0494            & 43.0677          & 0.9777          & 0.9667          & 13.1736           & 41.7560         & {\ul 0.9764}    & 0.5472          & 13.7319           \\
        Hysure            & 44.4228         & 0.9796         & 1.5854           & 13.6379            & 42.1455          & 0.9752          & 1.0637          & 15.2437           & 40.0435         & 0.9679          & 0.7116          & 19.0891           \\
        NLSTF             & 44.6407         & 0.9739         & 1.5569           & 15.8824            & 43.8243          & 0.9719          & {\ul 0.9022}    &  16.4862     & 43.0225         & 0.9708          & 0.5102          & {\ul 16.7782 }    \\
        CSTF              & {\ul45.1114}    & {\ul 0.9819}   & {\ul 1.5218}     & {\ul 9.7218 }     & {\ul 43.9305}    & {\ul 0.9779}    & 0.9100          & {\ul10.9041}           & {\ul 43.2382}   & 0.9763          & {\ul 0.4972}    & \textbf{11.7565}           \\
        CTDF              & 44.5731 	    & 0.9754 	     & 1.6875 	        & 15.4104  	       & 43.7258 	      & 0.9729          & 0.9664 	      & 16.0499  	        & 42.9958 	      & 0.9705 	        & 0.5471 	      & 16.5415            \\
        JSSO              & 44.4057 	    & 0.9766	     & 1.6515 	        & 14.8097  	       & 43.1283 	      & 0.9738          & 0.9779 	      & 15.4955  	        & 42.0972 	      & 0.9709 	        & 0.5533 	      & 16.2205            \\
        BGS-GTF           & 44.0998 	    & 0.9718 	     & 1.8255 	        & 14.7524  	       & 43.9275 	      & 0.9751          & 1.0540 	      & 13.4348  	        & 42.9904 	      & 0.9744 	        & 0.5957 	      & 12.9118            \\
        FCTN              & 39.9238 	    & 0.9422 	     & 2.7515 	        & 17.9608  	       & 39.3014 	      & 0.9371 	        & 1.5504 	      & 20.7791  	        & 39.8519 	      & 0.9538 	        & 0.7199 	      & 20.6506           \\[1pt]\hline
        BFCTN             & \textbf{45.5716} & \textbf{0.9839}& \textbf{1.5051}  & \textbf{9.1925 }  & \textbf{44.7692} & \textbf{0.9813} & \textbf{0.8246} & \textbf{10.2234 } & \textbf{43.4067}& \textbf{0.9768} & \textbf{0.4928} & {\ul11.8071}  
        \\[1pt]\hhline{=:=:=:=:=:=:=:=:=:=:=:=:=}
        \multirow{2}{*}{sf=4}     & \multicolumn{4}{c}{SNR=30dB}                                           & \multicolumn{4}{c}{SNR=20dB}                                           & \multicolumn{4}{c}{SNR=10dB}                                          
        \\ \cmidrule(lr){2-5}  \cmidrule(lr){6-9}  \cmidrule(lr){10-13} 
        & PSNR$\uparrow$  & SSIM$\uparrow$  & ERGAS$\downarrow$  & SAM $\downarrow$  & PSNR $\uparrow$   & SSIM$\uparrow$  & ERGAS$\downarrow$  & SAM$\downarrow$  & PSNR$\uparrow$  & SSIM $\uparrow$  & ERGAS$\downarrow$  & SAM$\downarrow$    \\ \hline
        LTTR             & 39.1668          & 0.9452          & 2.6493          & 20.6347           & 33.1139          & 0.7280          & 5.0356          & 32.9726           & 24.1907          & 0.2999         & 13.9589          & 43.2923          \\
        LTMR             & 39.3169          & 0.9491          & 2.5988          & 19.2410           & 33.8410          & 0.7522          & 4.6509          & 30.5259           & 25.2574          & 0.3222         & 12.4363          & 40.9665          \\
        GTNN             & 38.9318          & 0.9431          & 2.7207          & 19.8887           & 33.7008          & 0.7474          & 4.7345          & 30.6706           & 25.2776          & 0.3232         & 12.4152          & 41.1175          \\
        CNMF             & 42.2416          & 0.9620          & 2.0021          & 12.7530           & 35.3913          & 0.7920          & 3.8054          & 26.9579           & 25.5235          & 0.3677         & 12.2299          & 37.2262          \\
        Hysure           & 42.1788          & 0.9570          & 1.8998          & 18.5166           & 34.8497          & 0.7611          & 4.1139          & 29.9098           & 25.5975          & 0.3296         & 11.9135          & 40.4807          \\
        NLSTF            & 41.6080          & 0.9414          & 2.0341          & 21.7016           & 33.1815          & 0.7064          & 5.0047          & 33.7306           & 23.3800          & 0.2732         & 15.1707          & 43.9308          \\
        CSTF             & {\ul 43.7147}    & {\ul 0.9720}    & {\ul 1.7492}    & {\ul 12.7808 }    & 36.0514          & 0.8219          & 3.8833          & 28.8187           & 23.4527          & 0.2410         & 16.4888          & 44.0105          \\
        CTDF             & 41.7851 	        & 0.9454 	      & 2.1364 	        & 20.8440  	      & 33.8516 	     & 0.7293 	       & 4.7601 	     & 31.8729  	       & 24.7087 	      & 0.3081 	       & 13.3563 	      & 42.1607           \\
        JSSO             & 41.7999 	        & 0.9493 	      & 2.0458 	        & 19.9852  	       & 34.4676 	      & 0.7594          & 4.2996 	      & 30.4015  	        & 26.4122 	      & 0.3950 	        & 10.6123 	      & 38.6408            \\
        BGS-GTF          & 41.3763 	        & 0.9402 	      & 1.6875 	        & 21.7235  	       & 32.9561 	      & 0.6855          & 5.5753 	      & 33.1500  	        & 23.0509 	      & 0.2283 	        & 17.2621 	      & 43.3803            \\
        FCTN             & 39.6824          & 0.9411 	      & 2.8033 	        & 18.5512  	      &{\ul36.6623} 	 & {\ul0.8771}     & {\ul3.5900} 	 & {\ul24.4300 }     & {\ul29.3674} 	  & \textbf{0.5427}& {\ul7.6629} 	  & {\ul35.8467 }  \\[2pt]\hline
        BFCTN            & \textbf{44.2936} & \textbf{0.9767} & \textbf{1.6403} & \textbf{11.5959 } & \textbf{38.9409} &\textbf{0.8976}  & \textbf{2.6759} & \textbf{18.8010 } & \textbf{30.1761} & {\ul0.5421}    & \textbf{7.0628}  & \textbf{28.7665 }\\ [1pt] \hline
    \end{tabular}
    \label{tab:CAVE} 
\end{table*}

\begin{figure*} [h!]
    \includegraphics[width=\textwidth]{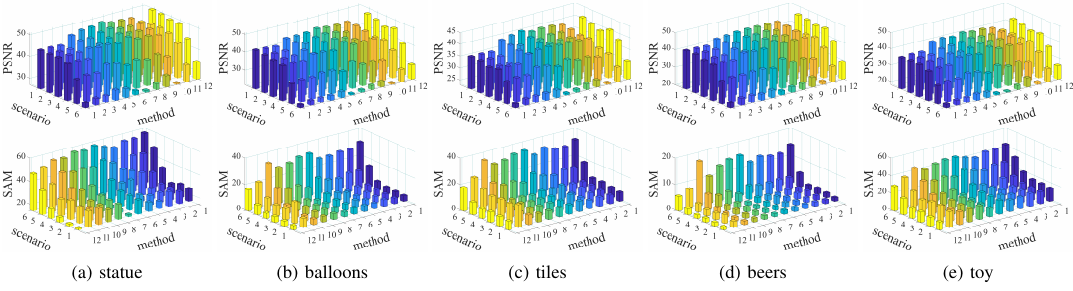}
    \caption{The PSNR and SAM of the five images in CAVE dataset are compared by different methods in different scenarios. Method 1-12 correspond to LTTR, LTMR, GTNN, CNMF, Hysure, NLSTF, CSTF, CTDF, JSSO, BGS-GTF, FCTN, and BFCTN, respectively.}
    \label{fig:bar3}
\end{figure*}

\subsection{Baseline and Quantitative Metrics}
In the field of HMF tasks, the effectiveness of methods needs to be verified by systematic comparisons across technology lines. The benchmark test set constructed in this study covers three types of mainstream paradigms: (1) the matrix decomposition methods are represented by CNMF~\cite{yokoya2011coupled} and Hysure~\cite{simoes2014convex}; (2) the tensor modeling system contains LTTR~\cite{dian2019learning}, LTMR~\cite{dian2019hyperspectral}, GTNN~\cite{dian2024hyperspectral}, NLSTF~\cite{dian2019nonlocal}, CSTF~\cite{li2018fusing}, CTDF~\cite{xu2024coupled}, JSSO~\cite{long2024joint}, BGS-GTF~\cite{wang2025generalized}, and FCTN~\cite{jin2022high}, which together explore tensor a priori representations with different rank constraints;  and (3) the deep learning system integrates emerging methods such as ZSL~\cite{dian2023zero}, M2U-Net~\cite{li2024model}, EDIP-Net~\cite{li2025enhanced}, LRTN~\cite{liu2025low}, reflecting the paradigm evolution from convolutional architectures to attention mechanisms. 

To establish an evaluation system simultaneously assessing spectral-spatial fidelity, we select four metrics: Peak Signal-to-Noise Ratio (PSNR)~\cite{dian2019hyperspectral}, Structural Similarity (SSIM)\cite{yokoya2017hyperspectral}, Erreur Relative Globale Adimensionnellede Synthèse (ERGAS)~\cite{wald2000quality}, and Spectral Angle Mapper (SAM)~\cite{yokoya2011coupled}, forming a multiscale framework from local to global levels.

\begin{figure*} []
    \centering
    \begin{subfigure}[c]{0.951\textwidth} 
        \centering
        \includegraphics[width=\linewidth]{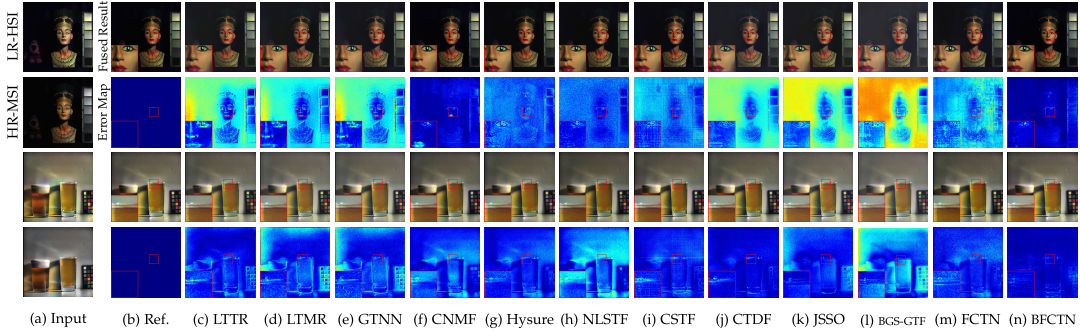} 
    \end{subfigure}
    \hfill
    \begin{subfigure}[c]{0.0404\textwidth} 
        \centering
        \raisebox{4.45mm}{\includegraphics[width=\linewidth]{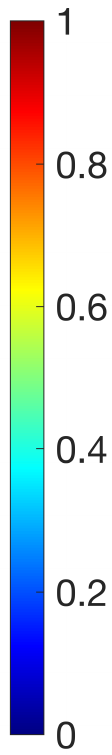}}
    \end{subfigure}

    \caption{Fusion results of other methods and BFCTN in Scenario1 on the 'statue' and 'beers' images in the CAVE dataset.}
    \label{fig:show_CAVE} 
\end{figure*}

\subsection{Datasets Description}
In this section, we introduce the three evaluation datasets and provide a detailed account of the process for generating synthetic LR-HSI and HR-MSI.

\textbf{CAVE Dataset}\footnote{\url{https://www.cs.columbia.edu/CAVE/databases/multispectral/}}: This dataset is acquired by an Apogee Alta U260 camera and consists of 32 indoor scenes captured by a composite pixel camera. Each image has dimensions of $512 \times 512 \times 31$ pixels, where $512 \times 512$ represents the spatial dimensions and 31 denotes the spectral bands, with wavelengths ranging from 400 nm to 700 nm at 10 nm intervals. The dataset encompasses a heterogeneous assortment of natural and artificial materials, objects, shapes, and colors. Five representative images—including statue, balloons, tiles, beers, and toy—are selected to evaluate all approaches.

\textbf{Harvard Dataset}\footnote{\url{http://vision.seas.harvard.edu/hyperspec/download.html}}: The dataset, provided by Harvard University, consists of 50 indoor and outdoor scenes captured under daylight illumination. Each image has dimensions of $1040 \times 1392 \times 31$, with spectral wavelengths spanning from 420 nm to 720 nm at 10 nm intervals. To facilitate a more accurate evaluation of all approaches, each image is cropped from the upper left corner to the size of $512 \times 512 \times 31$. Five representative images, including indoor and outdoor scenes—building (imga1), fruit (imgf8), car (imgc4), placard (imgd7), and wall (imge6)—are selected for evaluation.

\textbf{Pavia University Dataset}\footnote{\url{https://www.ehu.eus/ccwintco/index.php/}}: This dataset is acquired using the Reflective Optics System Imaging Spectrometer in Pavia, Italy, in 2003. The original dataset contains $610 \times 340 \times 115$ pixels, with wavelengths spanning from 430 nm to 860 nm. After excluding bands affected by water vapor interference and those with low signal-to-noise ratios, a subset of $256 \times 256 \times 91$ pixels is selected as the reference image.

The reference images are processed using the following approach to generate LR-HSI and HR-MSI. To generate LR-HSI, an averaging filter kernel of size $sf \times sf$ is initially applied to the reference images, followed by downsampling by a factor of $sf$. HR-MSI is generated by downsampling the hyperspectral dimensions based on the spectral response function specific to each image. The spectral response function is obtained from a Nikon D700 camera for both the CAVE and Harvard datasets. The 31 bands of the reference images are reduced to 3 bands, resulting in HR-MSI images that are analogous to traditional RGB images. In the Pavia dataset, IKONOS-like reflective spectral response filters are applied to simulate HR-MSI with dimensions of $256 \times 256 \times 4$. In order to create a realistic simulation, Gaussian noise at a specified level is added to the generated LR-HSI and HR-MSI.

Although the Wald protocol~\cite{wald1997fusion} seeks to guarantee impartiality in comparative testing, previous studies still suffer from inconsistent inputs, unclear parameters, and opaque experimental scenarios, etc. To address these issues, we extended the fairness protocol for comparing fusion methods, including but not limited to the following aspects:

1) To ensure the uniformity of all comparison experiments, all input images are generated using a consistent method.

2) All parameters of the comparison method are derived from the optimal settings described in the original paper without any additional optimisation or modification.

3) Comparison experiments are to be conducted under different downsampling factors and noise intensities with clearly defined experimental scenarios.
\subsection{Model Performance Evaluation}
\subsubsection{Experimental Setup}
For our method, the hyperparameters $a_0, b_0, c_0, d_0, e_0, f_0$ are empirically set to a small value of $1 \times 10^{-6}$. This choice serves as a non-informative prior and helps stabilize the inference process by avoiding the introduction of strong bias from the prior distribution. With the above, the initial values of $\tau_h$ and $\tau_m$ are to 1, and the parameters ${\boldsymbol\lambda^{k}_{n_1,n_2}}$ are initialized as all-one vectors. The initial factor tensors are randomly generated. 

For data preprocessing, the input HR-MSI is chunked into small patches of $64 \times 64$ pixels with an overlap of 48 pixels, while the corresponding LR-HSI is partitioned into patches of $64/sf \times 64/sf$, where $sf$ denotes the spatial downsampling factor. The maximum number of iterations is set to six throughout all experiments. The ranks of the BFCTN are set to the following values for the purposes of the fusion test: $[35, 4, 12, 4, 12, 4]$. The experiments encompass a range of scenarios, including different downsampling factor~($sf$) and $SNR$ levels, for a total of six scenarios: Scenario 1: $SNR$=35dB, $sf$=4; Scenario 2: $SNR$=35dB, $sf$=8; Scenario 3: $SNR$=35dB, $sf$=16; Scenario 4: $SNR$=30dB, $sf$=4; Scenario 5: $SNR$=20dB, $sf$=4; Scenario 6: $SNR$=10dB, $sf$=4.

\begin{table*}[ht!]
    \caption{The fusion performance of different methods in different scenarios is compared on the Harvard dataset.}
    \begin{tabular}{ccccccccccccc}
        \hline
        \multirow{2}{*}{$SNR$=35dB}              & \multicolumn{4}{c}{$sf$=4}                                             & \multicolumn{4}{c}{$sf$=8}                                              & \multicolumn{4}{c}{$sf$=16}                                             
        \\ \cmidrule(lr){2-5}  \cmidrule(lr){6-9}  \cmidrule(lr){10-13}  
        & PSNR$\uparrow$  & SSIM$\uparrow$  & ERGAS$\downarrow$  & SAM $\downarrow$  & PSNR $\uparrow$   & SSIM$\uparrow$  & ERGAS$\downarrow$  & SAM$\downarrow$  & PSNR$\uparrow$  & SSIM $\uparrow$  & ERGAS$\downarrow$  & SAM$\downarrow$    \\ \hline
        LTTR    & 41.9097       & 0.9733       & 2.6712         & 2.7895         & 42.2786       & 0.9745       & 1.6403         & 3.1136        & 42.4026          & 0.9747          & 0.9428            & 3.4836           \\
        LTMR    & 42.2513       & 0.9745       & 2.2957         & 2.7700         & 42.7363       & 0.9762       & 1.2254         & 3.1096        & 42.6472          & 0.9755          & 0.7095            & 3.5532           \\
        GTNN    & 41.9536       & 0.9711       & 2.3711         & 3.0533         & 42.6466       & 0.9758       & 1.2372         & 3.1343        & 42.7943          & 0.9764          & 0.6941            & 3.4566           \\
        CNMF    & 45.2396       & {\ul 0.9825} & 1.8666         & 2.4334         & 44.7029       & {\ul 0.9822} & 0.9764         & {\ul2.5801}   & 43.9750          & 0.9804          & 0.5108            & 2.6473           \\
        Hysure  & {\ul 45.5182} & 0.9818       & 2.1539         & {\ul 2.5345}   & 44.4464       & 0.9806       & 1.1151         & 2.9570        & 44.2252          & {\ul 0.9806}    & 0.5678            & 3.1677           \\
        NLSTF   & 45.2475       & 0.9798       & 2.2268         & 2.6322         & 44.9025       & 0.9795       & 1.1328         & 2.7315        & 44.6631          & 0.9796          & 0.5747            & 2.7910           \\
        CSTF    & 45.5062       & 0.9808       & {\ul 1.8356}   & 2.4953         & {\ul 45.2793} & 0.9805       & \textbf{0.9357}& 2.5765        & {\ul 45.0993}    & 0.9804          & \textbf{0.4775}  & \textbf{2.6410}     \\
        CTDF    & 45.2974       & 0.9782       & 2.0382         & 2.7195         & 44.9059       & 0.9776       & 1.0620         & 2.8350        & 44.5474          & 0.9769          & 0.5722            & 2.9624           \\
        JSSO    & 45.2707       & 0.9789       & 2.0817         & 3.2855         & 44.6178       & 0.9778       & 1.0724         & 2.8340        & 44.6178          & 0.9774          & 0.5480            & 2.9122           \\
        BGS-GTF  & 45.3475      & 0.9795       & 1.9316         & 2.6939         & 44.8203       & 0.9788       & 0.9856         & 2.7144        & 44.8203          & 0.9786          & 0.5091            & 2.8153           \\
        FCTN    & 42.8358       & 0.9691       & 2.3183         & 2.6440         & 43.7690       & 0.9749       & 1.1072         & 3.2457        & 41.1380          & 0.9540          & 1.0456            & 4.7776           \\ \hline
        BFCTN   & \textbf{45.8808} & \textbf{0.9830} & \textbf{1.8353}   & \textbf{2.3847}   & \textbf{45.6024} & \textbf{0.9826} & {\ul 0.9406}      & \textbf{2.4976}  & \textbf{45.1332} & \textbf{0.9815} & {\ul 0.5071}      & {\ul2.6425}  
        \\[1pt]\hhline{=:=:=:=:=:=:=:=:=:=:=:=:=}
        \multirow{2}{*}{$sf$=4}                 & \multicolumn{4}{c}{$SNR$=30dB}                                        & \multicolumn{4}{c}{$SNR$=20dB}                                         & \multicolumn{4}{c}{$SNR$=10dB}                                     
        \\ \cmidrule(lr){2-5}  \cmidrule(lr){6-9}  \cmidrule(lr){10-13}   
        & PSNR$\uparrow$  & SSIM$\uparrow$  & ERGAS$\downarrow$  & SAM $\downarrow$  & PSNR $\uparrow$   & SSIM$\uparrow$  & ERGAS$\downarrow$  & SAM$\downarrow$  & PSNR$\uparrow$  & SSIM $\uparrow$  & ERGAS$\downarrow$  & SAM$\downarrow$     \\ \hline 
        LTTR   & 40.5957        & 0.9607       & 3.3441         & 3.4718         & 34.7317       & 0.8284       & 7.5154          & 8.0674        & 26.0003        & 0.4336          & 20.6733           & 20.4302          \\
        LTMR   & 41.1664        & 0.9639       & 2.6832         & 3.3308         & 35.7396       & 0.8502       & 5.7702          & 7.1228        & 26.9195        & 0.4609          & 18.0715           & 18.5992          \\
        GTNN   & 40.8697        & 0.9598       & 2.7591         & 3.6596         & 35.5256       & 0.8434       & 5.9111          & 7.4447        & 26.9308        & 0.4616          & 18.0808           & 18.7175          \\
        CNMF   & 44.1168        & 0.9759       & 2.0595         & 2.8119         & 37.3961       & 0.8816       & 4.0419          & 6.4516        & 27.4888        & 0.5073          & 12.5086           & 17.8248          \\
        Hysure & 44.5412        & 0.9762       & 2.3096         & 2.7975         & 38.9690       & 0.9035       & 3.4328          & 5.0485        & 28.5060        & 0.5170          & 13.4146           & 16.0014          \\
        NLSTF  & 42.9151        & 0.9651       & 3.0635         & 3.4363         & 35.1294       & 0.8225       & 7.5372          & 8.3557        & 25.4193        & 0.4080          & 22.6059           & 21.6852          \\
        CSTF   & {\ul 45.1177}  & {\ul 0.9787} & {\ul 1.8830}   & {\ul 2.5851}   & 38.8764       & 0.9078       & 3.7190          & 5.4473        & 25.4159        & 0.3770          & 22.7173           & 22.6101          \\
        CTDF   & 43.1215        & 0.9632       & 2.6949         & 3.5153         & 35.5798       & 0.8283       & 6.7639          & 8.1149        & 26.2244        & 0.4355          & 18.0391           & 20.7401          \\
        JSSO   & 43.3163        & 0.9659       & 2.6504         & 3.4057         & 36.9816       & 0.8747       & 5.5484          & 6.6828        & 28.9617        & 0.5796          & 13.3913           & 15.0167          \\
        BGS-GTF & 43.9173       & 0.9715       & 2.2391         & 3.1782         & 36.1647       & 0.8482       & 5.9833          & 7.8325        & 25.1843        & 0.3750          & 22.7556           & 23.5266          \\
        FCTN   & 42.6357        & 0.9672       & 2.3450         & 3.2866         & {\ul 40.6183} & {\ul 0.9413} & {\ul 2.7292}    & {\ul 3.9614}  & {\ul 33.0894}  & {\ul 0.7254}    & \textbf{6.0059}   & {\ul 8.9635}     \\ [1pt] \hline
        BFCTN  & \textbf{45.4224} & \textbf{0.9808} & \textbf{1.8791}   & \textbf{2.4667}   & \textbf{41.8917} & \textbf{0.9517} & \textbf{2.3993}   & \textbf{3.2292}  & \textbf{33.2077} & \textbf{0.7379} & {\ul 6.4861}      & \textbf{8.1082}  \\ [1pt] \hline
    \end{tabular}
    \label{tab:Harvard} 
\end{table*}

\subsubsection{Experimental on CAVE Dataset}
A comprehensive experimental investigation is conducted on five experimental images of the CAVE dataset, encompassing six scenarios in the experimental setting. The objective is to evaluate and compare the performance of eleven traditional comparison methods alongside the proposed BFCTN. The numerical evaluation results presented in Tab. \ref{tab:CAVE} are obtained by averaging the indicators of the five images, facilitating an intuitive assessment of the performance of the various methods on the CAVE dataset. As evidenced in Tab. \ref{tab:CAVE}, the BFCTN exhibits superior performance compared to other comparison methods across all resolutions, particularly in the context of Scenario 2, where its indicators significantly surpass those of all comparison methods. Furthermore, methods such as CNMF, CSTF, and Hysure demonstrate high performance under high $SNR$ conditions, whereas methods such as LTTR and LTMR exhibit relatively poor performance. As the $SNR$ decreases to 30 dB, 20 dB, and 10 dB, the BFCTN continues to demonstrate superior performance across all indicators. Furthermore, CSTF and FCTN demonstrate robustness under medium $SNR$ conditions, whereas the performance of other comparison methods, except BFCTN, declines considerably at lower $SNR$ levels. Fig. \ref{fig:bar3} provides a visual representation of the comprehensive fusion performance of different methods for each image under all input scenarios, clearly showing that BFCTN exhibits excellent fusion performance across various images and all $SNR$ and downsampling factor conditions.

Fig. \ref{fig:show_CAVE} presents the output and error maps for two images with contrasting backgrounds in Scenario 1: a darker image of a 'statue' and a brighter image of 'beers'. The initial row of each set of images presents a false color composite image synthesized by bands 28, 16, and 1, while the subsequent row illustrates the error image between the fusion result and the reference image. Fig. \ref{fig:show_CAVE}~(a) depicts the LR-HSR and HR-MSI used for fusion. Fig. \ref{fig:show_CAVE}~(b) shows the reference image, and Figs. \ref{fig:show_CAVE}~(c)-(n) show the fusion results of the eleven comparison methods and proposed BFCTN.

Overall, all experimental methods effectively extract high-resolution details from the HR-MSI and fuse them with the LR-HSI to generate satisfactory HR-HSI. However, from the locally magnified details and error images, in the 'statue' image with a darker background, the LTTR, LTMR, GTNN, CTDF, JSSO and BGS-GTF exhibit significant discrepancies from the reference image in the error maps. The CNMF performs well overall in fusion but has noticeable errors in the locally magnified portrait area. The Hysure and NLSTF introduce a significant amount of noise after fusion, while the CSTF and FCTN cause severe distortion in the darker background areas. For the 'beers' image with a brighter background, the fusion effect of the various methods improves in the background. However, the LTMR, GTNN, and NLSTF still show distortions at the background edges. Additionally, the LTTR, CNMF, CSTF, CTDF and BGS-GTF show pronounced streaks and artifacts around the cup edge in the magnified area, while the FCTN still exhibits a considerable degree of noise. In contrast, BFCTN demonstrates superior clarity in the false-color images, with sharper edges and smoother error maps.

\subsubsection{Experimental on Harvard Dataset}
In addition, we also conduct comparative experiments in six scenarios on five images of the Harvard dataset. Tab. \ref{tab:Harvard} illustrates the assessment outcomes on the Harvard dataset, wherein the proposed BFCTN exhibits optimal performance across a range of metrics in diverse scenarios. It is worthy of note that in Scenarios 2, 3 and 6, the optimal ERGAS values are achieved by CSTF and FCTN, respectively. This phenomenon can be attributed to the potential for significant errors in specific bands or local areas of the fused image. Furthermore, it is notable that under conditions of high noise, the performance of FCTN markedly surpasses that of other comparison methods, approaching the performance of the proposed BFCTN.

\begin{figure*} [t!]
    \centering
    \begin{subfigure}[c]{0.951\textwidth} 
        \centering
        \includegraphics[width=\linewidth]{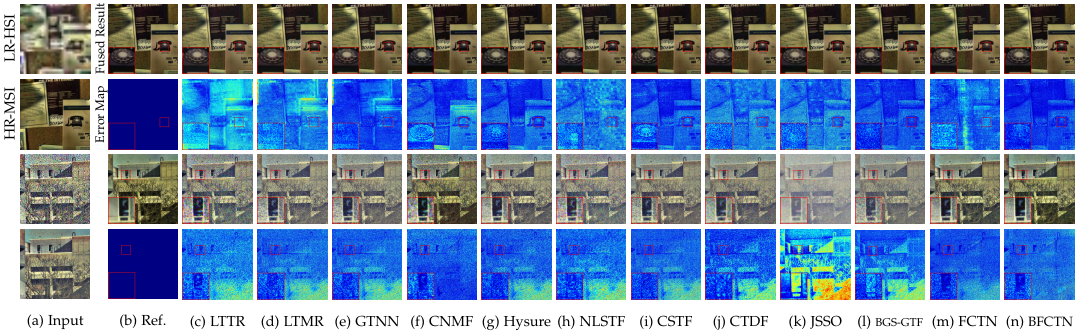} 
    \end{subfigure}
    \hfill
    \begin{subfigure}[c]{0.0404\textwidth} 
        \centering
        \raisebox{4.45mm}{\includegraphics[width=\linewidth]{colorbar.pdf}}
    \end{subfigure}
    
    \caption{Fusion results of other methods and proposed BFCTN on the 'placard' and 'building' images in the Harvard dataset. The ’placard‘ image is experimented with in Scenario 3, while building is experimented with in Scenario 6.}
    \label{fig:show_Harvard} 
\end{figure*}

To facilitate a more intuitive comparison of the performance of comparison methods alongside the proposed BFCTN under two low-quality input scenarios, we select one indoor image 'placard' and one outdoor image 'building' for display in Fig.~\ref{fig:show_Harvard}. The 'placard' is experimented with in Scenario~3, while the 'building' is experimented with in Scenario~6. As can be observed in Fig. \ref{fig:show_Harvard}~(a), strong blur and high noise are clearly evident in the 'placard' and 'building', respectively. Unlike the CAVE dataset, false-color images for the Harvard dataset are synthesized using bands 20, 15, and 8. Upon examination of the fusion results for the 'placard', it becomes evident that even with sparser sampling, satisfactory outcomes can still be achieved. However, it is notable that there are discernible differences in the details observed among the methods. The LTTR, LTMR, CTDF and JSSO exhibit significant color distortion, and the GTNN shows numerous streaks in the error map. The fusion images from CNMF, Hysure, and NLSTF retain a significant amount of noise. In contrast, the CSTF and FCTN produce images with high clarity, but they also exhibit significant errors in darker background areas. In the case of the 'building' image, higher noise levels are applied, resulting in the majority of methods retaining a considerable amount of noise in their fusion results. Only the CNMF, FCTN, and BFCTN demonstrate satisfactory performance. However, in the locally magnified areas of the false-color image, the first two exhibit severe blurring, while the BFCTN maintains clear details and sharp edges.

\begin{figure} [t!]
    \centering
    \includegraphics[width=\linewidth]{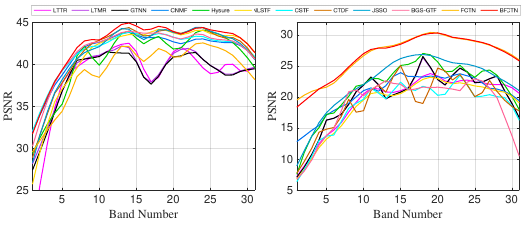}

    \caption{The PSNR of each fusion result of 'placard' and 'building' along the spectral bands}
    \label{fig:Harvard_rmse_psnr} 
\end{figure}

Furthermore, the distinct bands are subjected to visual analysis with regard to the two aforementioned images, as illustrated in Fig. \ref{fig:Harvard_rmse_psnr}. In the highly blurred 'placard' image, the proposed BFCTN achieves favorable results, with smooth lines~(red) maintaining the highest PSNR in most locations. In the 'building' image affected by strong noise interference, both the FCTN and BFCTN significantly outperform other methods. The advantage of BFCTN over FCTN is relatively minor, indicating that the FCTN also exhibits robust performance.

\subsection{Deep Learning Baseline Comparison} 

\begin{table}[b!]
\centering
\caption{Comparison results with advanced deep learning methods.}

\begin{tabular}{
    >{\centering}p{0.6cm} | 
    >{\centering}p{0.7cm} 
    c 
    >{\centering}p{0.8cm} 
    >{\centering}p{0.8cm} 
    >{\centering}p{0.8cm} 
    c
}
\hline
$SNR$                    & Dataset                  & Method   & PSNR↑            & SSIM↑           & ERGAS↓          & SAM↓              \\ \hline
\multirow{10}{*}{30dB} & \multirow{5}{*}{CAVE}    & ZSL      & 34.2167          & 0.9063          & 1.5111          & 19.6241          \\
                       &                          & M2U-Net  & 39.4336          & 0.9419          & 1.0719          & {\ul 12.6855}    \\
                       &                          & EDIP-Net & {\ul 39.7096}    & {\ul 0.9445}    & {\ul 0.9868}    & \textbf{12.5048} \\
                       &                          & LRTN     & 38.7608          & 0.9404          & 1.0911          & 15.7430          \\
                       &                          & BFCTN    & \textbf{39.8753} & \textbf{0.9517} & \textbf{0.9711} & 14.9580          \\ \cline{2-7} 
                       & \multirow{5}{*}{Harvard} & ZSL      & 40.6948          & 0.9577          & 8.4937          & 7.8108           \\
                       &                          & M2U-Net  & 46.1575          & 0.9823          & {\ul 2.2935}    & {\ul 4.4542}     \\
                       &                          & EDIP-Net & {\ul 47.3979}    & {\ul 0.9854}    & 2.4114          & 4.8031           \\
                       &                          & LRTN     & 41.6276          & 0.9721          & 3.1580          & 7.5513           \\
                       &                          & BFCTN    & \textbf{47.9209} & \textbf{0.9877} & \textbf{1.8129} & \textbf{4.2952}  \\ \hline
\multirow{10}{*}{10dB} & \multirow{5}{*}{CAVE}    & ZSL      & 20.0787          & 0.2096          & 8.4254          & 43.0520          \\
                       &                          & M2U-Net  & 29.7534          & {\ul0.8388}          & 2.6536          & {\ul 23.4079}    \\
                       &                          & EDIP-Net & {\ul 30.0779}    & \textbf{0.8413}          & {\ul 2.5507}    & 23.6223          \\
                       &                          & LRTN     & 22.4170          & 0.3264          & 6.2037          & 38.1593          \\
                       &                          & BFCTN    & \textbf{30.4776} & 0.7299          & \textbf{2.4536} & \textbf{23.0515} \\ \cline{2-7} 
                       & \multirow{5}{*}{Harvard} & ZSL      & 27.2621          & 0.5439          & 17.2224         & 29.6534          \\
                       &                          & M2U-Net  & 34.9699          & {\ul 0.8825}    & 5.4545          & 10.0532          \\
                       &                          & EDIP-Net & {\ul 35.0370}    & 0.8814          & {\ul 5.4160}    & {\ul 10.0521}    \\
                       &                          & LRTN     & 27.9139          & 0.5125          & 22.5027         & 34.4219          \\
                       &                          & BFCTN    & \textbf{41.2420} & \textbf{0.9450} & \textbf{3.3745} & \textbf{7.2982}  \\ \hline
\end{tabular}
\label{deep}
\end{table}

The aforementioned analysis proved the effectiveness of BFCTN against conventional methods. However, recent advances in the field of deep learning have introduced the latest paradigm of hyperspectral multispectral image fusion. In order to comprehensively position our proposed method in the contemporary research field, we extend our comparative analysis to four representative state-of-the-art deep learning methods: ZSL\cite{dian2023zero}, M2U-Net\cite{li2024model}, EDIP-Net\cite{li2025enhanced}, and LRTN~\cite{liu2025low}. These methods are selected for their established spectral-spatial reconstruction performance and open-source implementations. Specifically, we respectively selected the CAVE ('toy') and Harvard ('imgf8') datasets. Building on previous observations that fusion results exhibit low sensitivity to the downsampling factor~($sf$), we fix sf = 8 in this section to focus on evaluating performance under varying additive noise levels, consistent with established protocols.

Tab.~\ref{deep} shows the quantitative results, under low noise scenario~(30dB), BFCTN achieves the highest PSNR, SSIM, and ERGAS scores on the CAVE dataset, though it slightly underperforms in SAM compared to ZSL and M2U-Net. On the Harvard dataset, BFCTN consistently outperforms all four deep learning baselines across all metrics, highlighting its robustness in more complex scenarios. In high-noise scenario~(10dB), BFCTN maintains leading performance in PSNR, ERGAS, and SAM for the CAVE dataset, while exhibiting a marginal deficit in SSIM against EDIP-Net and LRTN. Notably, it secures comprehensive superiority on the Harvard dataset, even under severe noise degradation. These results suggest that while deep learning methods excel in certain spectral or structural metrics (e.g., SAM or SSIM), BFCTN provides more reliable and generalizable reconstruction quality, particularly in challenging noise environments. 
The unsatisfactory performance of ZSL and LRTN under high-noise conditions mainly arises from their sensitivity to distributional shifts. Although ZSL aims to generalize without retraining, its reconstruction heavily depends on feature similarity learned from clean training distributions, leading to degraded estimation when the spectral observations are heavily corrupted. Likewise, the deterministic low-rank constraint in LRTN lacks explicit noise modeling, making it less robust to severe perturbations.

\begin{figure} [t!]
    \includegraphics[width=\linewidth]{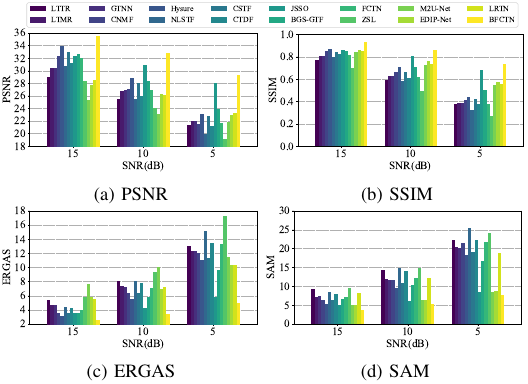}
    \caption{Properties of different methods for Pavia University datasets under high noise conditions.}
    \label{fig:PU_bar} 
\end{figure}

\begin{figure*} [b!]
    \centering
    \begin{subfigure}[c]{0.978\textwidth} 
        \centering
        \includegraphics[width=\linewidth]{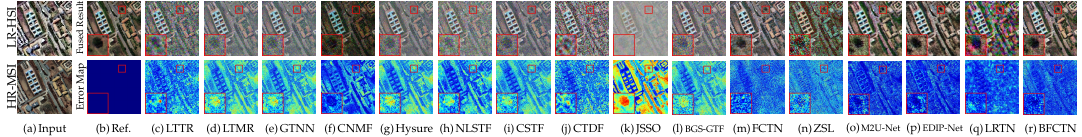} 
    \end{subfigure}
    \hfill
    \begin{subfigure}[c]{0.016\textwidth} 
        \centering
        \raisebox{3mm}{\includegraphics[width=\linewidth]{colorbar.pdf}}
    \end{subfigure}
    \caption{ Fusion results of other methods and BFCTN on the Pavia University dataset. Input images are subjected to Gaussian white noise with $SNR$ = 5dB, and the LR-HSI is downsampled by a factor of $sf$ = 4.}
    \label{fig:show_PU} 
\end{figure*}

\subsection{Model Robustness Evaluation} 

In previous experiments, the proposed BFCTN consistently delivers exceptional performance across a wide range of scenarios. It is noteworthy that both BFCTN and FCTN demonstrate superior performance compared to other methods in low $SNR$ scenarios. To further evaluate the noise immunity of FCTN in comparison to the proposed BFCTN and test the robustness of various methods, comparative experiments are conducted with special emphasis on extremely low $SNR$ scenarios. Given that the dimensions of the Pavia University dataset following cropping are $256 \times 256$, the ranks of BFCTN are adjusted to [35, 4, 3, 4, 3, 2] to align with these dimensions. The parameters of other compared methods are set in accordance with the original paper's specifications. For some methods lacking explicit Pavia University dataset parameters, the previous parameters are utilized in the experiments.

Fig. \ref{fig:PU_bar} illustrates the performance of eleven comparison methods and the proposed BFCTN on the Pavia University $SNR$ levels of 15, 10, and 5 dB, as presented in bar charts. The yellow bars represent the proposed BFCTN, which is observed to outperform other methods in numerical terms. Moreover, FCTN, which is also based on FCTN decomposition, exhibits robust performance under conditions of extremely low~$SNR$.

Fig. \ref{fig:show_PU} illustrates the fused images obtained using various methods at an $SNR$ of 5 dB. The initial row presents false-color images synthesized from bands 42, 15, and 8. Although CNMF is effective in reducing noise, it introduces the most conspicuous color distortion. In contrast to the other methods, BFCTN not only eliminates noise but also preserves image details with exceptional accuracy. In the magnified regions, only FCTN, M2U-Net, EDIP-Net and BFCTN effectively restore the shrub details. However, the FCTN produces images with blurring and artifacts in other areas, and some noise from M2U-Net and EDIP-Net still remains. The second row displays the error images of the fused results in comparison to the ground truth. It is evident that BFCTN produces darker and smoother error images. In the presence of significant noise interference, the fusion performance of BFCTN is demonstrably superior to that of other methods.

\subsection{Evaluation on Real Dataset} 
To validate the practical applicability of the proposed method, we conducted experiments on a real-world dataset acquired by the Ziyuan(ZY)-02D satellite. The dataset contains spatially co-registered LR-HSI, HR-MSI, and panchromatic (PAN) data. The original LR-HSI has dimensions of 6213$\times$6099$\times$166 with spectral coverage from 395 to 2501 nm, while the HR-MSI measures 2071$\times$2033$\times$8 bands. We removed long-wavelength bands with relatively low signal-to-noise ratios and selected 76 spectral bands covering visible and near-infrared regions. A 800$\times$800$\times$76 subregion is cropped and downsampled as the LR-HSI, with a corresponding 100$\times$100$\times$8 subregion extracted from the HR-MSI. Based on the CNMF degradation modeling framework, we improved the estimation process of the spectral response function (SRF). Specifically, on the basis of nonnegative constraints, a normalization constraint (ensuring that the sum of response coefficients in each channel equals one) and a spectral smoothness regularization term were introduced to avoid unreasonable oscillations between adjacent bands. At the same time, a weighting mechanism was applied to enhance robustness to invalid pixels. Using the updated SRF, the point spread function (PSF) was further estimated to enhance the stability of spatial degradation modeling, making the obtained degradation kernels more consistent with the physical characteristics of the imaging system and providing reliable priors for subsequent fusion.

\begin{figure*} [t!]
    \centering
    \includegraphics[width=\textwidth]{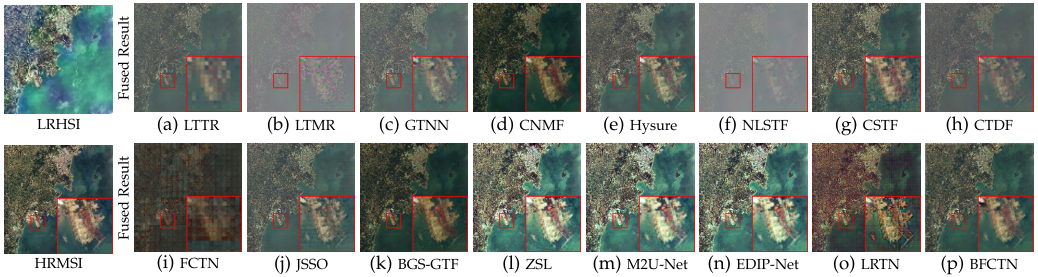}
    \caption{Fusion results of all comparison methods and BFCTN on real datasets.}
    \label{fig:show_real} 
\end{figure*}

\begin{figure} [h!]
    \centering
    \includegraphics[width=0.95\linewidth]{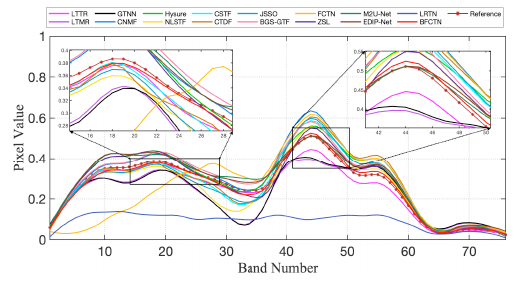}
    \caption{The spectral characteristic curves of all methods and input LR-HSI at [120,120] pixel positions.}
    \label{fig:pixel_ziyuan} 
\end{figure}

\begin{table*}[t!]
\centering
\caption{Comparison of quantitative indices (NHQM↓) of different methods on the real data.}
\setlength{\tabcolsep}{4pt}
\renewcommand{\arraystretch}{1.0}
\begin{adjustbox}{max width=\textwidth}
\begin{tabular}{cccccccccccccccc}
\hline
LTTR  & LTMR  & GTNN  & CNMF  & Hysure & NLSTF & CSTF  & CTDF  & FCTN  & JSSO  & BGS-GTF & ZSL   & M2U-Net & EDIP-Net & LRTN  & BFCTN \\ \hline
28.76 & 26.96 & 27.27 & 26.05 & 25.53  & 27.08 & 24.69 & 28.67 & 29.55 & 25.52 & 27.43   & 26.54 & 26.42   & 26.54    & 27.57 & 24.44 \\ \hline
\end{tabular}
\end{adjustbox}
\label{nhqm}
\end{table*}

\begin{table*}[t!]
    \caption{Average running time(seconds) of different methods.}
    \centering
    \begin{tabular}{ccccccccccccc}
    \hline
    Datasets         & LTTR   & LTMR   & GTNN   & CNMF  & Hysure & NLSTF & CSTF   & CTDF   & JSSO  & BGS-GTF & FCTN    & BFCTN   \\ \hline
    CAVE             & 352.05 & 230.10 & 243.45 & 38.70 & 127.17 & 69.71 & 119.84 & 48.24  & 5.55  & 178.42  & 3251.52 & 849.98  \\
    Harvard          & 348.12 & 256.97 & 273.52 & 50.07 & 144.08 & 77.16 & 135.37 & 76.95  & 4.31  & 149.27  & 2963.19 & 922.94  \\
    Pavia University & 78.65  & 154.82 & 75.41  & 7.89  & 22.32  & 26.04 & 85.99  & 11.08  & 1.87  & 106.81  & 890.27  & 89.01   \\
    Real Dataset     & 457.76 & 362.96 & 381.27 & 58.05 & 198.53 & 96.08 & 244.39 & 116.67 & 12.55 & 232.52  & 3499.43 & 1321.44 \\ \hline
    \end{tabular}
    \label{time}
\end{table*}

In the absence of ground-truth references, the evaluation relied on the physical plausibility of the spatial texture and the consistency of the spectral features with the input LR-HSI. Visually, as shown in Fig.~\ref{fig:show_real}, BFCTN achieves an optimal balance between spatial enhancement and spectral preservation, as evidenced by minimized artifacts, spatial clarity close to HR-MSI, and the absence of spectral aberrations in vegetated areas. Furthermore, Tab.~\ref{nhqm} reports the no-reference hyperspectral quality metric (NHQM)\cite{tian2022sparse} results, where BFCTN achieves the highest score, further confirming its superior capability improve spatial details on real data. In addition, Fig.~\ref{fig:pixel_ziyuan} demonstrates the comparison of the spectral features of the outputs of each method with the input LR-HSI at a random pixel, and the outputs of CSTF, EDIP-Net and BFCTN maintain a high degree of consistency with the spectral features of the input image. The above comparison shows that BFCTN has excellent ability to fuse real datasets.

\subsection{Time Overhead Analysis}
Tab.~\ref{time} summarizes the time requirements for each method. Due to the training requirements of deep learning methods, this comparison is not made. The proposed BFCTN requires multiple high-order tensor decompositions, which is a more time-consuming process than methods based on matrix decomposition and sparse tensor decomposition. However, our method is significantly faster than FCTN, which is based on the same decomposition paradigm. This is because FCTN raises the data to the 5-th order, introducing greater computational overhead with increasing dimensions, slowing down the convergence of the model. Thus, BFCTN strikes an effective balance between time cost and fusion performance.

\subsection{Discussion}
This section discusses BFCTN from the perspectives of architectural effectiveness and implementation, including the contribution of the Gamma prior, FCTN-ranks selection, chunking and overlapping strategies, the impact of various spatial degradation kernels, and the model convergence analysis. All experiments are conducted using the 'toy' image from the CAVE dataset as the test subject.

\subsubsection{Contribution of Hierarchical Sparsity Priors} 
To verify the validity of the hierarchical sparsity prior, we compared the original model with Gamma distribution and the variant with fixed sparse constraints. The test group replaced the Gamma distributions of the top-level parameters $\tau_h^k$, $\tau_m^k$ and $\boldsymbol{\lambda}^k_{(n_1,n_2)} $ with constants ($C$ $\in$  $\left\{1e-2,1e-1,1\right\}$), and left the rest of the model settings unchanged. The experiments are executed with three parameter configurations ($\boldsymbol{\lambda} $ fixed (include $\boldsymbol{\lambda}^k_{(n_1,n_2)} $), $\tau$ fixed (include $\tau_h^k$ and $\tau_m^k$),  and full parameter fixed) are tested, and the performance metrics on the dataset are recorded.

Tab.~\ref{Tab:Gamma} shows the quantitative results, where the $\boldsymbol{\lambda} $ shows good robustness (fluctuations of $<$ 0.4 dB in the range of 1e-2 to 1), while the $\tau$ exhibits strong sensitivity (6.2 dB decrease in PSNR when fixed to 1e-2 and only 0.1 dB when fixed to 1). It is noteworthy that when both $\boldsymbol{\lambda}$ and $\tau$ are fixed to constants, the results seem to possess the properties of the above two parameter configurations, with stable but lower results than the original configuration when fixed at smaller values, and performance closer to the original configuration when fixed at 1. These results not only verify the validity of the hierarchical Gamma prior mechanism, but also show its certain ability to be parameter free and provide more stable results.

\subsubsection{Effect of FCTN-ranks}
\begin{figure} [t!]
    \centering
    \includegraphics[width=\linewidth]{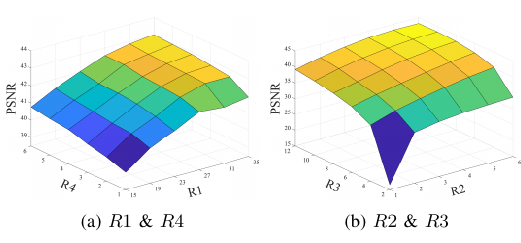}
    \caption{The rank selection strategy in the proposed method.}
    \label{fig:rank_analyze} 
\end{figure}

In Eq.~\eqref{FCTN}, the FCTN-ranks consist of a set of numbers $(R_{1,2}, \dots, R_{1,N}, \dots, R_{N-1,N})$. In the proposed model, given that the tensor to be decomposed has the same scale in the first and second dimensions, it is reasonable to consider $R_{1,3}$ and $R_{2,3}$ to be equivalent, as well as $R_{1,4}$ and $R_{2,4}$. Then the parameters of the rank are restricted to four, i.e., we use $R1$ to denote $R_{1,2}$, $R2$ to denote $R_{1,3}$ and $R_{2,3}$, $R3$ to denote $R_{1,4}$ and $R_{2,4}$, and $R4$ to denote $R_{3,4}$. Further, we categorize them into two groups and discuss the synergistic influence relationship between $R1$, $R4$ and $R2$, $R3$, respectively. As Fig.~\ref{fig:rank_analyze}~(a) illustrates, larger $R1$ values result in better performance, while improvements with $R4$ diminish beyond 4. Balancing time cost and performance, we select $R1$ as 35 and $R4$ as 4. Additionally, from Fig.~\ref{fig:rank_analyze}(b), we observe a slight performance drop for $R2$ beyond 4, while $R3$ shows steady improvement. Thus, we set $R2$ to 4 and $R3$ to 12. The final FCTN-ranks are determined to be [35, 4, 12, 4, 12, 4].

\begin{table}[t!]
\caption{Comparison between Gamma prior and fixed parameters.}
\centering
\begin{tabular}{cc|cccc}
\hline
\multicolumn{2}{c|}{Configuration}                             & PSNR↑            & SSIM↑           & ERGAS↓          & SAM↓             \\ \hline
\multicolumn{2}{c|}{Original (Gamma)}                          & \textbf{42.7628} & \textbf{0.9764} & \textbf{1.4199} & \textbf{10.5293} \\ \hline
\multirow{3}{*}{$\boldsymbol{\lambda} $ fixed}          & 1e-2 & 41.4363          & 0.9735          & 1.6196          & 10.8248          \\
                                                        & 1e-1 & 41.4164          & 0.9735          & 1.6321          & 11.8382          \\
                                                        & 1    & 41.8382          & 0.9761          & 1.5178          & 10.8832          \\ \hline

\multirow{3}{*}{$\tau$ fixed}                           & 1e-2 & 36.5949          & 0.8920          & 2.5775          & 12.2121          \\
                                                        & 1e-1 & 41.7180          & 0.9744          & 1.5010          & 10.6793          \\
                                                        & 1    & 42.6246          & 0.9745          & 1.4578          & 10.5929          \\ \hline
\multirow{3}{*}{$\boldsymbol{\lambda}$ \& $\tau$   fixed} & 1e-2 & 41.3683          & 0.9732          & 1.6250          & 10.8621          \\
                                                        & 1e-1 & 41.3771          & 0.9733          & 1.6213          & 10.9152          \\
                                                        & 1    & 42.2795          & 0.9728          & 1.5315          & 10.7180          \\ \hline
\end{tabular}
\label{Tab:Gamma}
\end{table}

\subsubsection{Effect of Chunking and Overlapping}
\begin{figure} [h!]
    \centering
    \includegraphics[width=\linewidth]{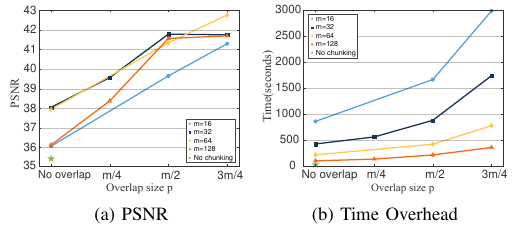}
    \caption{Analysis of performance and time overhead by different chunking and overlapping strategies.}
    \label{fig:mp} 
\end{figure}

To optimize the chunking and overlapping strategies parameters, we designed controlled experiments. The system evaluates the combination of patch size $m \in \left\{16,32,64,128\right\}$ with overlapping region $p$ (grouping constraint: $\frac{W-p}{m-p} + 1 = K$). As shown in Fig.~\ref{fig:mp}~(a), the optimal PSNR is obtained for $m$=64 with $p$=48 ($3m/4 $ overlap), which is a 1.0 dB improvement over the suboptimal combination ($m$=32, $p$=16). When $m$=512 is set (i.e., the full graph input is not chunked), the model degrades to a third-order tensor network structure. The experimental data show that the model performance metrics drop significantly in this case, and this comparison result strongly validates the design effectiveness of the chunking strategy.

Fig.~\ref{fig:mp}~(b) demonstrates the effect of different parameters on the time overhead, it is worth mentioning that the different combinations directly affect the $K$-value, the model is processed in groups, and the different group number $K$ do not affect the model performance, but its effect on the time performance is larger. Combining performance and computational cost, the model selects a patch size of 64 and an overlap size of 48.

\subsubsection{Degradation Kernels Comparison}

To systematically investigate the impact of kernel separability assumptions on fusion performance, we design a controlled experiment with four representative degradation kernels, which include:

\begin{itemize}
    \item \textbf{Average kernel}: $4 \times 4$ average kernel.
    \item \textbf{Gaussian kernel $G_{4\times4}$}: $\sigma = 2$ with $4 \times 4$ support.
    \item \textbf{Gaussian kernel $G_{7\times7}$}: $\sigma = 2$ with $7 \times 7$ support.
    \item \textbf{Motion blur kernel $M_{30^\circ}$}: length = $4$ and $\theta = 30^\circ$.
    \item \textbf{Motion blur kernel $M_{45^\circ}$}: length = $4$ and $\theta = 45^\circ$.
    \item \textbf{Elliptical kernel $E_{30^\circ}$}: Anisotropic Gaussian blur with $\sigma_x = 1$, $\sigma_y = 3$, and rotation angle $\theta = 30^\circ$, simulating non-isotropic optical defocus.
    \item \textbf{Hybrid kernel $H_{G+M}$}: Convolution of a Gaussian kernel ($\sigma = 2$) and a motion blur kernel (length = $4$, $\theta = 30^\circ$), representing compound degradations combining defocus and motion effects.
    \item \textbf{Sensor-specific kernel $S_{\text{var}}$}: Channel-dependent Gaussian blurs with increasing $\sigma$ across spectral bands, modeling heterogeneous spatial responses of multi-sensor imaging systems.
\end{itemize}

Where the average and Gaussian kernels can be exactly decomposed into the outer product of 1D vectors and are therefore regarded as separable kernels, the motion blur, elliptical, hybrid, and sensor-varying kernels are non-separable due to their directional or spatially heterogeneous characteristics. For methods that require the decomposition of the point spread function (PSF), such non-separable kernels are approximated by performing an SVD-based low-rank factorization after zero-padding.

\begin{table}[h!]
\centering
\caption{The performance of BFCTN under different degradation kernels.}
\begin{tabular}{ccccc}
\hline
Kernel  & PSNR↑            & SSIM↑           & ERGAS↓          & SAM↓   \\ \hline
Average               & 42.7628 & 0.9764 & 1.4199 & 10.5293 \\
$G_{4\times4}$        & 42.9082 & 0.9760 & 1.3915 & 10.4246 \\
$G_{7\times7}$        & 42.2129 & 0.9743 & 1.4976 & 10.8569 \\
$M_{30^\circ}$        & 41.4712 & 0.9723 & 1.5288 & 10.4926 \\
$M_{45^\circ}$        & 41.3125 & 0.9718 & 1.5531 & 10.6378 \\
$E_{30^\circ}$        & 40.9856 & 0.9702 & 1.6024 & 10.9845 \\
$H_{G+M}$             & 41.2728 & 0.9765 & 1.5733 & 10.7319 \\
$S_{\text{var}}$      & 40.8651 & 0.9669 & 1.7440 & 11.5923 \\ \hline
\end{tabular}
\label{kernel} 
\end{table}

\begin{table}[h!]
\centering
\caption{The performance of various methods under the motion blur kernel.}
\begin{tabular}{ccccc}
\hline
Method  & PSNR↑            & SSIM↑           & ERGAS↓          & SAM↓                                   \\ \hline
LTTR    & 35.9105                              & 0.9534                              & 2.5996                              & 14.4999                              \\
LTMR    & 36.1927                              & 0.9556                              & 2.5026                              & 13.7503                              \\
GTNN    & 35.8465                              & 0.9486                              & 2.6096                              & 15.1712                              \\
CNMF    & 38.8291                              & 0.9558                              & 2.0359                              & 13.6203                              \\
Hysure  & 40.8093                              & 0.9594                              & 1.8441                              & 14.8049                              \\
NLSTF   & 34.2629                              & 0.9361                              & 3.1578                              & 16.2967                              \\
CSTF    & 38.2946                              & {\ul 0.9660}                        & 2.1279                              & {\ul 12.2264}                        \\
CTDF    & 37.0001                              & 0.9106                              & 2.3732                              & 19.5667                              \\
JSSO    & 35.4651                              & 0.9038                              & 3.0848                              & 19.6212                              \\
BGS-GTF & {\ul 41.0662}                        & 0.9536                              & {\ul 1.7415}                        & 15.0933                              \\
FCTN    & 40.7569                              & 0.9474                              & 1.9461                              & 14.7243                              \\ \hline
BFCTN   & {\textbf{41.4712}} & {\textbf{0.9723}} & {\textbf{1.5288}} & {\textbf{10.4926}} \\ \hline
\end{tabular}
\label{motion} 
\end{table}

As shown in Tab.~\ref{kernel}, the PSNR difference among the three standard separable kernels (Average, $G_{4\times4}$, and $G_{7\times7}$) is less than $0.7$~dB, indicating consistent high performance under standard blurring. In contrast, the motion blur kernels ($M_{30^\circ}$ and $M_{45^\circ}$) lead to a PSNR drop of more than $1$~dB due to directional information loss during the SVD approximation. Notably, BFCTN outperforms competing methods under the motion blur condition (Tab.~\ref{motion}), achieving a $20\%$ relative improvement in SAM (from $12.23^\circ$ to $10.49^\circ$), it surpasses the fusion method BGS-GTF applicable to general spatial blurring. For the newly introduced complex kernels, including the elliptical Gaussian kernel ($E_{30^\circ}$), the hybrid Gaussian-motion kernel ($H_{G+M}$), and the channel-varying sensor-specific kernel ($S_{\text{var}}$), only BFCTN results are reported due to computational constraints. Despite the increased difficulty of these degradations, BFCTN still achieves reasonable performance, demonstrating its robustness and adaptability. These results suggest that BFCTN can effectively handle a variety of separable and non-separable, directional, and heterogeneous degradations.

\subsubsection{Convergence Analysis}
\begin{figure} [t!]
    \centering
    \includegraphics[width=\linewidth]{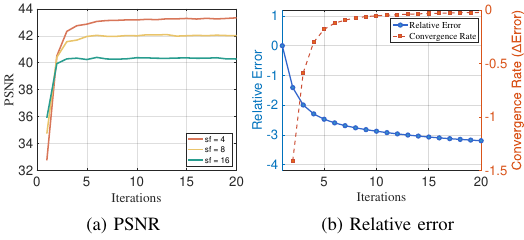}
    \caption{The convergence behavior of the proposed method.}
    \label{fig:convergence} 
\end{figure}

Fig. \ref{fig:convergence}~(a) illustrates the PSNR convergence process of the model under different downsampling factors. As the number of iterations increases, the PSNR gradually stabilizes. Fig. \ref{fig:convergence}~(b) further analyzes the convergence characteristics at a downsampling factor of $sf$=4, where the relative error\footnote{Relative error = $\log \left( \|\mathcal{X}^{t+1}-\mathcal{X}^{t}\|_F / \|\mathcal{X}^{t}\|_F \right)$, where $\mathcal{X}^{t}$ and $\mathcal{X}^{t+1}$ represent the previous iteration and the current result respectively. \cite{zhou2019bayesian}.} and error convergence rate curves demonstrate that the error reduction diminishes significantly after five iterations, while the PSNR curve reaches a plateau, Both visual quality and numerical stability indicate that additional iterations yield negligible performance improvement. Based on these observations, the maximum number of iterations is empirically set to six in our implementation to balance efficiency and performance.

\section{Conclusion}
\label{sec:conclusion}
This paper presents a hyperspectral-multispectral image fusio model, designated BFCTN, which is constructed on the foundation of Bayesian learning and the FCTN decomposition paradigm. By employing chunking and overlapping techniques, the model is able to efficiently extract the low-rank properties of the input images. Subsequently, We construct a fully connected tensor network structure by setting up a hierarchical sparse prior between factor tensors, which enables the model to take full advantage of the intrinsic structure of the data and the correlation between different dimensions. Furthermore, we develop a method for estimating model parameters using VB and EM, which effectively avoids the non-convex optimisation problem common to other fusion methods and greatly reduces the complexity of parameter selection. Extensive comparative experiments across multiple datasets and diverse input scenarios, the proposed BFCTN method demonstrates not only state-of-the-art fusion performance and remarkable robustness against strong noise interference and various degradation effects, but also demonstrates practical applicability on real datasets.

\ifCLASSOPTIONcompsoc

\else
\fi

\ifCLASSOPTIONcaptionsoff
\newpage
\fi

\bibliographystyle{IEEEtran}

\cleardoublepage
\onecolumn
\appendices
\appendix

\subsection{The posterior estimation of factor tensors $\mathcal{T}^{k}_{n}$.}
\label{appendix: factor}
\subsubsection{The posterior estimation of $\mathcal{T}^{k}_{1}$}
The estimation of $\mathcal{T}^{k}_{1}$ is presented in Sec. \ref{sec:model}. B of the paper (main document). 
	
\subsubsection{The posterior estimation of $\mathcal{T}^{k}_{2}$}
	
For $\mathcal{T}^{k}_{2}$, the optimization objective function is
\begin{equation*}
	\begin{aligned}
		& \max \mathbb{E}_{q\left(\boldsymbol{\Xi}^{k} \backslash \mathcal{T}^{k}_2\right)}
		\Bigg\{ -\frac{\tau_h^{k}}{2}\left\|\mathcal {H}^{k}-\mathbb{FCTN}\left ( \mathcal{T}^{k}_1\times_1 \mathbf{P}_1,\mathcal{T}^{k}_2\times_2 \mathbf{P}_2,\mathcal{T}^{k}_3 ,\mathcal{T}^{k}_4\right)\right\|_{\mathrm{F}}^2\\
		& \quad -\frac{\tau_m^{k}}{2}\left\|\mathcal{M}^{k}-\mathbb{FCTN}\left ( \mathcal{T}^{k}_1,\mathcal{T}^{k}_2,\mathcal{T}^{k}_3 \times_3 \mathbf{P}_3,\mathcal{T}^{k}_4\right) \right\|_{\mathrm{F}}^2
		-\frac{1}{2} \operatorname{Tr}\left(\mathbf{T}^{k}_{2(2)} \mathbf{\Lambda}^{k}_{2}{\mathbf{T}^{k}_{2(2)}}^{\top}\right)\Bigg\}\\    
	\end{aligned}
\end{equation*}
According to the EM rule, optimizing the above formula can obtain a Sylvette equation as follows:
	
\begin{equation*}
	\begin{aligned}
		&\mathbf{T}^{k}_{2(2)}\left\{\mathbb{E}\left[\tau^{k}_{m}\right] \mathbb{E}\left[{\mathbf{M}^{k}_{\neq2}}^\top\mathbf{M}^{k}_{\neq2}\right]+\mathbb{E}\left[\boldsymbol{\Lambda}^{k}_2\right]\right\}+\mathbb{E}\left[\tau^{k}_{h}\right] \mathbf{P_2}^{\top} \mathbf{P_2} \mathbf{T}^{k}_{2(2)} \mathbb{E}\left[{\mathbf{H}^{k}_{\neq2}}^{\top}\mathbf{H}^{k}_{\neq2}\right] 
		\\
        &\qquad=\mathbb{E}\left[\tau_{m}^{k}\right]\mathbf{M}^{k}_{(2)} \mathbb{E}\left[{\mathbf{M}^{k}_{\neq2}}^{\top}\right]+\mathbb{E}\left[\tau_{h}^{k}\right] \mathbf{P_2}^{\top}\mathbf{H}^{k}_{(2)} \mathbb{E}\left[{\mathbf{H}^{k}_{\neq2}}^{\top}\right] . \\
	\end{aligned}
\end{equation*}	
Here, $\mathbf{T}^{k}_{2(2)}$ is regarded as the $X$ variable, and solving the Sylvette equation of the form $\mathbf{A} \mathbf{X} \mathbf{B}+\mathbf{C} \mathbf{X} \mathbf{D}=\mathbf{E}$ can yield the estimated value of $\mathbf{T}^{k}_{2(2)}$. Further folding of $\mathbf{T}^{k}_{2(2)}$ is the estimated value of $\mathcal{T}^{k}_{2}$.
	
\subsubsection{The posterior estimation of $\mathcal{T}^{k}_{3}$}
	
For $\mathcal{T}^{k}_{3}$, the optimization problem of the model is
\begin{equation*}
	\begin{aligned}
		& \max \mathbb{E}_{q\left(\boldsymbol{\Xi}^{k} \backslash \mathcal{T}^{k}_3\right)}\Bigg\{ -\frac{\tau_h^{k}}{2}\left\|\mathcal {H}^{k}-\mathbb{FCTN}\left ( \mathcal{T}^{k}_1\times_1 \mathbf{P}_1,\mathcal{T}^{k}_2\times_2 \mathbf{P}_2,\mathcal{T}^{k}_3 ,\mathcal{T}^{k}_4\right)\right\|_{\mathrm{F}}^2\\
		& \quad -\frac{\tau_m^{k}}{2}\left\|\mathcal{M}^{k}-\mathbb{FCTN}\left ( \mathcal{T}^{k}_1,\mathcal{T}^{k}_2,\mathcal{T}^{k}_3 \times_3 \mathbf{P}_3,\mathcal{T}^{k}_4\right) \right\|_{\mathrm{F}}^2
		-\frac{1}{2} \operatorname{Tr}\left(\mathbf{T}^{k}_{3(3)} \mathbf{\Lambda}^{k}_{3}{\mathbf{T}^{k}_{3(3)}}^{\top}\right)\Bigg\}\\    
	\end{aligned}
\end{equation*}
Similarly, the estimation of $\mathcal{T}^{k}_{3}$ can be obtained by solving the following linear system
\begin{equation*}
	\begin{aligned}
		&\mathbf{T}^k_{3(3)}\left\{\mathbb{E}\left[\tau^k_{h}\right] \mathbb{E}\left[{\mathbf{H}^k_{\neq3}}^{\top}\mathbf{H}^k_{\neq3}\right]+\mathbb{E}\left[\boldsymbol{\Lambda}^k_3\right]\right\}+\mathbb{E}\left[\tau^k_{m}\right] \mathbf{P_3}^{\top} \mathbf{P_3} \mathbf{T}^k_{3(3)} \mathbb{E}\left[{\mathbf{M}^k_{\neq3}}^{\top}\mathbf{M}^k_{\neq3}\right] 
		\\
        &\qquad=\mathbb{E}\left[\tau^k_m\right]\mathbf{P_3}^{\top}\mathbf{M}^k_{(3)} \mathbb{E}\left[{\mathbf{M}^k_{\neq3}}^{\top}\right]+\mathbb{E}\left[\tau^k_h\right] \mathbf{H}^k_{(3)} \mathbb{E}\left[{\mathbf{H}^k_{\neq3}}^{\top}\right] . \\
	\end{aligned}
\end{equation*}
	
\subsubsection{The posterior estimation of $\mathcal{T}^{k}_{4}$}
	
For $\mathcal{T}^{k}_{4}$, the optimization problem of the model is
\begin{equation*}
	\begin{aligned}
		& \max \mathbb{E}_{q\left(\boldsymbol{\Xi}^{k} \backslash \mathcal{T}^{k}_4\right)}
		\Bigg\{ -\frac{\tau_h^{k}}{2}\left\|\mathcal {H}^{k}-\mathbb{FCTN}\left ( \mathcal{T}^{k}_1\times_1 \mathbf{P}_1,\mathcal{T}^{k}_2\times_2 \mathbf{P}_2,\mathcal{T}^{k}_3 ,\mathcal{T}^{k}_4\right)\right\|_{\mathrm{F}}^2\\
		& \quad -\frac{\tau_m^{k}}{2}\left\|\mathcal{M}^{k}-\mathbb{FCTN}\left ( \mathcal{T}^{k}_1,\mathcal{T}^{k}_2,\mathcal{T}^{k}_3 \times_3 \mathbf{P}_3,\mathcal{T}^{k}_4\right) \right\|_{\mathrm{F}}^2
		-\frac{1}{2} \operatorname{Tr}\left(\mathbf{T}^{k}_{4(4)} \mathbf{\Lambda}^{k}_{4}{\mathbf{T}^{k}_{4(4)}}^{\top}\right)\Bigg\}\\    
	\end{aligned}
\end{equation*}
Similarly, the estimation of $\mathcal{T}^{k}_{4}$ can be obtained by solving the following linear system
\begin{equation*}
	\begin{aligned}
		& \mathbf{T}^k_{4(4)} \mathbb{E}\left\{\left[\tau^k_h\right] \mathbb{E}\left[{\mathbf{H}^k_{\neq4}}^{\top}\mathbf{H}^k_{\neq4}\right] +\mathbb{E}\left[\tau^k_m\right] \mathbb{E}\left[{\mathbf{M}^k_{\neq4}}^{\top}\mathbf{M}^k_{\neq4}\right]+\mathbb{E}\left[\boldsymbol{\Lambda}^k_4\right]\right\} 
		=\mathbb{E}\left[\tau^k_h\right]\mathbf{H}^k_{(4)} \mathbb{E}\left[{\mathbf{H}^k_{\neq4}}^{\top}\right]+\mathbb{E}\left[\tau^k_m\right] \mathbf{M}^k_{(4)} \mathbb{E}\left[{\mathbf{M}^k_{\neq4}}^{\top}\right] . \\
	\end{aligned}
\end{equation*}

\subsection{The posterior distribution of $\boldsymbol{\lambda}^{(n_1, n_2)}$}
\label{appendix: lambda}
\begin{equation*}
	\begin{aligned}
		q\left ( \boldsymbol{\lambda}^k_{(n_1,n_2)}  \right ) 
		&\propto \exp \left\{\mathbb{E}_{q(\boldsymbol{\Xi}-\boldsymbol{\lambda}^k_{(n_1,n_2)})}\left[\ln\mathcal{L}\left ( \mathbf{T}^k_{n_1},\mathbf{T}^k_{n_2} \mid \boldsymbol{\Lambda}^k_{n_1},\boldsymbol{\Lambda}^k_{n_2} \right ) + \ln p\left ( \boldsymbol\lambda^k_{(n_1,n_2)} \mid a_0,b_0 \right ) \right]\right\} \\
		&\propto \prod_{r}^{R_{n_1,n_2}} \exp \Bigg\{ \mathbb{E}_{q}\Bigg [ \frac{1}{2} \ln\left |  \boldsymbol{\Lambda}^k_{n_1} \right | - \frac{1}{2} {\mathbf{T}^k_{n_1({n_2})}(r)}^\top\boldsymbol{\Lambda}^k_{n_1} \mathbf{T}^k_{n_1(n_2)}(r)  \Bigg.\\ \Bigg.   &\quad+ \frac{1}{2} \ln\left |  \boldsymbol{\Lambda}^k_{n_2} \right | -\frac{1}{2}{\mathbf{T}^k_{n_2(n_1)}(r)}^{\top}\boldsymbol{\Lambda}^k_{n_2} \mathbf{T}^k_{n_2(n_1)}(r) + (a_0 - 1)\ln(\lambda^k_{n_1,n_2})_{r} -b_0(\lambda^k_{n_1,n_2})_{r}\Bigg ] \Bigg\} \\
		&\propto \prod_{r}^{R_{n_1,n_2}} \exp \Bigg\{ \mathbb{E}_{q}\Bigg [ \left ( a_0+\frac{I_{n_1}\prod_{\substack{n\neq{n_1,n_2}}}^{N}R_{n,n_1} + I_{n_2}\prod_{\substack{n\neq{n_1,n_2}}}^{N}R_{n,n_2} }{2} -1 \right )\ln(\lambda^k_{n_1,n_2})_{r} \big.\\ \big.&\quad -\left ( b_0 + \frac{1}{2}\left (  \boldsymbol{\lambda}^{k}_{n_1,\neq n_2} {\mathbf{T}^k_{n_1(n_2)}(r) }^2 + \boldsymbol{\lambda}^{k}_{n_2,\neq n_1} {\mathbf{T}^k_{n_2(n_1)}(r) }^2 \right ) \right )(\lambda^k_{n_1,n_2})_{r} \Bigg ] \Bigg\} \\
		&\propto \prod_{r}^{R_{n_1,n_2}} {(\lambda^k_{n_1,n_2})_r}^{a_0+\frac{I_{n_1}\prod_{\substack{n\neq{n_1,n_2}}}^{N}R_{n,n_1} + I_{n_2}\prod_{\substack{n\neq{n_1,n_2}}}^{N}R_{n,n_2} }{2} - 1} \\ 
            & \quad \cdot exp\left \{  -\left(  b_0 + \frac{1}{2}\left (  \mathbb{E}_{q}\left[{\mathbf{T}^{k}_{n_1(n_2)}(r) }^{\top}\boldsymbol{\lambda}^{k}_{n_1,\neq n_2} {\mathbf{T}^{k}_{n_1(n_2)}(r) } \right]\right. + \mathbb{E}_{q}\left[{\mathbf{T}^{k}_{n_2(n_1)}(r)}^{\top}\boldsymbol{\lambda}^{k}_{n_2,\neq n_1} {\mathbf{T}^{k}_{n_2(n_1)}(r) } \right]\right ){(\lambda^k_{n_1,n_2})_{r}} \right\} \\
		&\propto \prod_{r}^{R_{n_1,n_2}} \text{Ga}\left((\lambda^{k}_{n_1, n_2})_r\mid a^{k}_{n_1,n_2}, b^{k}_{n_1,n_2}\right) 
	\end{aligned}
\end{equation*}
Therefore, the posterior of $\boldsymbol{\lambda}^{(n_1, n_2)}$ also follows the gamma distribution, and the posterior parameters can be obtained from the above form, where\\
\begin{equation*}
	\begin{aligned}
		& a^{k}_{n_1,n_2}=a_0+\frac{I_{n_1}\prod_{\substack{n\neq{n_1,n_2}}}^{N}R_{n,n_1} + I_{n_2}\prod_{\substack{n\neq{n_1,n_2}}}^{N}R_{n,n_2} }{2}\\
		& b^{k}_{n_1,n_2}=b_0 + \frac{1}{2}\left (  \mathbb{E}_{q}\left[{\mathbf{T}^{k}_{n_1(n_2)}(r) }^{\top}\boldsymbol{\lambda}^{k}_{n_1,\neq n_2} {\mathbf{T}^{k}_{n_1(n_2)}(r) } \right] + \mathbb{E}_{q}\left[{\mathbf{T}^{k}_{n_2(n_1)}(r)}^{\top}\boldsymbol{\lambda}^{k}_{n_2,\neq n_1} {\mathbf{T}^{k}_{n_2(n_1)}(r) } \right]\right ).
	\end{aligned}
\end{equation*}
	
\subsection{The posterior distribution of $\tau^k_h$}
\label{appendix: tau_h}
	
\begin{equation*}
	\begin{aligned}
		q \left( \tau^k_h \right ) 
		& \propto \exp \left\{\mathbb{E}_{q(\boldsymbol{\Xi}-\tau^k_h)}\left[\ln \mathcal{L}\left({\mathcal{H}}^k \mid\left\{\mathcal{T}^k_n\right\}_{n=1}^4, \tau^k_h\right)+\ln p(\tau^k_h)\right]\right\} \\
		& \propto \exp \left\{\mathbb{E}_{q}\left [\ln\left ( {\tau^k_h}^{\frac{I'_1I'_2I_3I_4}{2} } \right ) -\frac{\tau^k_h}{2}\sum_{i'_1}^{I'_1} \sum_{i_2}^{I'_2} \sum_{i_3}^{I_3}\sum_{i_4}^{I_4} \left ( \mathcal{H}^k_{i'_1i'_2i_3i_4}-\hat{\mathcal{H}}^k_{i'_1i'_2i_3i_4} \right )^2   + \ln\left ({\tau^k_h}^{c_0 - 1} \right ) - d_0\tau^k_h \right ] \right\} \\
		& \propto {\tau^k_h}^{c_0 + \frac{I'_1I'_2I_3I_4}{2} - 1}\exp \left\{ -\left (   d_0+\frac{1}{2}\mathbb{E}_{q}\left [ \left \| \left ( {\mathcal{H}}^k - \hat{{\mathcal{H}}}^k \right )   \right \|_F^2 \right ] \right )\tau^k_h \right\} \\
		& \propto {\tau^k_h}^{c_0 + \frac{I'_1I'_2I_3I_4}{2} - 1}\exp \left\{ -\left (   d_0+\frac{1}{2}\mathbb{E}_{q}\left [ \left \| \left ( {\mathcal{H}}^k - \mathbb{FCTN}\left ( {\mathcal{T}}^k_1\times_1 \mathbf{P}_1,{\mathcal{T}}^k_2\times_2 \mathbf{P}_2,{\mathcal{T}}^k_3 ,{\mathcal{T}}^k_4 \right) \right )   \right \|_F^2 \right ] \right )\tau^k_h \right\} \\
		& \propto Ga\left(\tau^k_h \mid c^k,d^k \right)
	\end{aligned}
\end{equation*}	
Thus, the posterior of $\tau^k_h$ is a parametric updated Gamma distribution, where the parameters are updated according to the following rule:
\begin{equation*}
	\begin{aligned}
		& c^{k}=c_0+\frac{I'_1 I'_2 I_3 I_4}{2} \\
		& d^{k}=d_0+ \frac{1}{2} \mathbb{E}_q\left[\left\|\left({\mathcal {H}^{k}}-\mathbb{FCTN}\left(\mathcal{T}^{k}_1 \times_1 \mathbf{P}_1, \mathcal {T}^{k}_2 \times_2 \mathbf{P}_2, \mathcal{T}^{k}_3, \mathcal{T}^{k}_4\right)\right)\right\|_F^2\right].
	\end{aligned}
\end{equation*}

\subsection{The posterior distribution of $\tau^k_m$}
\label{appendix: tau_m}
\begin{equation*}
	\begin{aligned}
		q \left( \tau^k_m \right ) 
		& \propto \exp \left\{\mathbb{E}_{q(\boldsymbol{\Xi}-\tau^k_m)}\left[\ln \mathcal{L}\left({\mathcal{M}}^k \mid\left\{\mathcal{T}^k_n\right\}_{n=1}^4, \tau^k_m\right)+\ln p(\tau^k_m)\right]\right\} \\
		& \propto \exp \left\{\mathbb{E}_{q}\left [\ln\left ( {\tau^k_m}^{\frac{I_1I_2I'_3I_4}{2} } \right ) -\frac{\tau^k_m}{2}\sum_{i_1}^{I_1} \sum_{i_2}^{I_2} \sum_{i'_3}^{I'_3}\sum_{i_4}^{I_4} \left ( \mathcal{M}^k_{i_1i_2i'_3i_4}-\hat{\mathcal{M}}^k_{i_1i_2i'_3i_4} \right )^2   + \ln\left ({\tau^k_m}^{e_0 - 1} \right ) - f_0\tau^k_m \right ] \right\} \\
		& \propto {\tau^k_m}^{e_0 + \frac{I_1I_2I'_3I_4}{2} - 1}\exp \left\{ -\left (   f_0+\frac{1}{2}\mathbb{E}_{q}\left [ \left \| \left ( {\mathcal{M}}^k - \hat{{\mathcal{M}}}^k \right )   \right \|_F^2 \right ] \right )\tau^k_m \right\} \\
		& \propto {\tau^k_m}^{e_0 + \frac{I_1I_2J_3I_4}{2} - 1}\exp \left\{ -\left (   f_0+\frac{1}{2}\mathbb{E}_{q}\left [ \left \| \left ( {\mathcal{M}}^k - \mathbb{FCTN}\left ( {\mathcal{T}}^k_1,{\mathcal{T}}^k_2,{\mathcal{T}}^k_3 \times_3 \mathbf{P}_3,{\mathcal{T}}^k_4 \right) \right )   \right \|_F^2 \right ] \right )\tau^k_m \right\} \\
		& \propto Ga\left(\tau^k_m \mid e^k,f^k \right)
	\end{aligned}
\end{equation*}	
Similarly, the posterior of $\tau^k_m$ is a parametric updated Gamma distribution, where the parameters are updated according to the following rule:
\begin{equation*}
	\begin{aligned}
		& e^{k}=e_0+\frac{I_1 I_2 I'_3 I_4}{2} \\
		& f^{k}=f_0+\frac{1}{2} \mathbb{E}_q\left[\left\|\left({\mathcal { M }^{k} }-\mathbb{FCTN}\left(\mathcal{T}^{k}_1, \mathcal {T}^{k}_2, \mathcal{T}^{k}_3 \times_3 \mathbf{P}_3, \mathcal{T}^{k}_4\right)\right)\right\|_F^2\right].
	\end{aligned}
\end{equation*}

\subsection{Lower bound of model evidence}
\label{appendix: ELBO}
\begin{equation*}
\begin{aligned}
        ELBO & =\mathbb{E}_{q(\Xi^k )}\left[\ln p\left(\mathcal{H}^k,\mathcal{M}^k, \Xi^k \right)\right]+H(q(\Xi^k )) \\
        & =\mathbb{E}_{q\left(\left\{\mathcal{T}^{(k)}_n\right\}^4_{n=1}, \tau^k_{h}\right)}\left[\ln p\left({\mathcal { H }}^k \mid\left\{\mathcal{T}^k_{n}\right\}^{4}_{n=1}, \tau^k_{h}\right)\right]+\mathbb{E}_{q\left(\left\{\mathcal{T}^{(k)}_n\right\}^4_{n=1}, \tau^k_{m}\right)}\left[\ln p\left({\mathcal { M }}^k \mid\left\{\mathcal{T}^k_{n}\right\}^{4}_{n=1}, \tau^k_{m}\right)\right] \\
        &+ \mathbb{E}_{q\left(\left\{\mathcal{T}^k_{n}\right\}^4_{n=1}\right)}\left[\sum_{n=1}^{4} \ln p\left(\mathcal{T}^{k}_{n} \mid \lambda^{k}_{1,n},\dots, \lambda^{k}_{n-1,n},\lambda^{k}_{n,n+1},\dots, \lambda^{k}_{n,N}\right)\right]+\mathbb{E}_{q(\boldsymbol{\lambda^k_{n_1,n_2}})}\left [   \sum_{n_1=1}^{3}\sum_{n_2=n_1+1}^{4}  \ln p\left(\boldsymbol\lambda^{k}_{n_1, n_2}\right)\right ] \\
        & +\mathbb{E}_{q(\tau^{k}_{h})}[\ln p\left(\tau^{k}_{h}\right)] + \mathbb{E}_{q(\tau^{k}_{m})}[\ln p\left(\tau^{k}_{m}\right)]-\mathbb{E}_{q\left(\left\{\mathcal{T}^{(k)}_n\right\}^4_{n=1}\right)}\left[\sum_{n=1}^{4} \ln q\left(\mathcal{T}^{(k)}_n\right)\right]-\mathbb{E}_{q(\boldsymbol\lambda^{k}_{n_1, n_2})}\left[\sum_{n_1=1}^{3}\sum_{n_2=n_1+1}^{4}  \ln q\left(\boldsymbol\lambda^{k}_{n_1, n_2}\right)\right]\\
        &-\mathbb{E}_{q(\tau^k_h)}[\ln q(\tau^k_h)] - \mathbb{E}_{q(\tau^k_m)}[\ln q(\tau^k_m)]
\end{aligned}
\end{equation*}	

The evidence lower bound of model consists of $\mathbb{E}_{q(\Xi^k )}\left[\ln p\left(\mathcal{H}^k,\mathcal{M}^k, \Xi^k \right)\right]$ and $H(q(\Xi^k ))$, the former is the expectation of the log prior distribution of the parameter and the latter is the entropy of the posterior distribution of the parameter, where:

\begin{equation*}
\begin{aligned}
\mathbb{E}_{q} & {\left[\ln p\left(\mathcal{H}^k \mid\left\{\mathcal{T}^k_{(n)}\right\}^4_{n=1}, \tau^{k}_h\right)\right] } \\
& =-\frac{I_1'I_2'I_3I_4}{2} \left(\ln (2 \pi)-\mathbb{E}_{q}[\ln \tau^k_h]\right)-\frac{1}{2} \mathbb{E}_{q}[\tau^k_h] \mathbb{E}_{q}\left[\left\|\mathcal {H}^{k}-\mathbb{FCTN}\left ( \mathcal{T}^{k}_1\times_1 \mathbf{P}_1,\mathcal{T}^{k}_2\times_2 \mathbf{P}_2,\mathcal{T}^{k}_3 ,\mathcal{T}^{k}_4\right)\right\|_{F}^{2}\right] \\
& =\frac{I_1'I_2'I_3I_4}{2}\left(\psi\left(c^k\right)-\ln d^k\right)-\frac{c^k}{2 d^k} \mathbb{E}_{q}\left[\left\|\mathcal {H}^{k}-\mathbb{FCTN}\left ( \mathcal{T}^{k}_1\times_1 \mathbf{P}_1,\mathcal{T}^{k}_2\times_2 \mathbf{P}_2,\mathcal{T}^{k}_3 ,\mathcal{T}^{k}_4\right)\right\|_{F}^{2}\right] + C
\end{aligned}
\end{equation*}	

\begin{equation*}
\begin{aligned}
\mathbb{E}_{q} & {\left[\ln p\left(\mathcal{M}^k \mid\left\{\mathcal{T}^k_{(n)}\right\}^4_{n=1}, \tau^{k}_m\right)\right] } \\
& =-\frac{I_1I_2I_3'I_4}{2} \left(\ln (2 \pi)-\frac{1}{2}\mathbb{E}_{q}[\ln \tau^k_m]\right)-\frac{1}{2} \mathbb{E}_{q}[\tau^k_m] \mathbb{E}_{q}\left[\left\|\mathcal {M}^{k}-\mathbb{FCTN}\left (\mathcal{T}^{k}_1,\mathcal{T}^{k}_2,\mathcal{T}^{k}_3 \times_3 \mathbf{P}_3,\mathcal{T}^{k}_4\right) \right\|_{\mathrm{F}}^2\right] \\
& =\frac{I_1I_2I_3'I_4}{2}\left(\psi\left(e^k\right)-\ln f^k\right)-\frac{e^k}{2 f^k} \mathbb{E}_{q}\left[\left\|\mathcal {M}^{k}-\mathbb{FCTN}\left (\mathcal{T}^{k}_1,\mathcal{T}^{k}_2,\mathcal{T}^{k}_3 \times_3 \mathbf{P}_3,\mathcal{T}^{k}_4\right) \right\|_{\mathrm{F}}^2\right] + C
\end{aligned}
\end{equation*}	

\begin{equation*}
\begin{aligned}
 & \mathbb{E}_q\left[\sum_{n=1}^4\ln p\left(\mathcal{T}^k_{n}\mid \boldsymbol\lambda^{k}_{1,n},\dots, \boldsymbol\lambda^{k}_{n-1,n},\boldsymbol\lambda^{k}_{n,n+1},\dots, \boldsymbol\lambda^{k}_{n,N}\right)\right] \\
 & =\mathbb{E}_q\left[\sum_{n=1}^{4}\sum^{I_n}_{i_n}\left\{\frac{1}{2}\ln\left | \boldsymbol{\Lambda}^{k}_{n} \right |  -\frac{1}{2}{\mathbf{T}^{k}_{n}(i_n) }^{\top}\mathbf{\Lambda}^{k}_{n} {\mathbf{T}^{k}_{n}(i_n) }\right\}\right] + C \\
 & =\sum_{n=1}^{4}\left\{\frac{I_n}{2}\left(\sum_r^{R_{1,n}}\mathbb{E}_q[\ln\lambda_{1,n}^k] +\cdots +\sum_r^{R_{n-1,n}}\mathbb{E}_q[\ln\lambda_{n-1,n}^k] + \cdots + \sum_r^{R_{n,4}}\mathbb{E}_q[\ln\lambda_{n,4}^k]\right)-\frac{1}{2}\sum_{i_n}^{I_n}\mathbb{E}_q\left[\mathbf{T}^{k}_{n} (i_n)^{\top}\mathbf{\Lambda}^{k}_{n}\mathbf{T}^{k}_{n}(i_n)\right]\right\}  + C\\
 & =\frac{\sum_nI_n}{2}\left(\sum_r^{R_{1,n}}\mathbb{E}_q[\ln\lambda_{1,n}^k] +\cdots +\sum_r^{R_{n-1,n}}\mathbb{E}_q[\ln\lambda_{n-1,n}^k] + \cdots + \sum_r^{R_{n,4}}\mathbb{E}_q[\ln\lambda_{n,4}^k]\right)\\
 &\quad-\frac{1}{2}\sum_{n=1}^{4}\sum_{i_n}^{I_n}\left\{\mathrm{Tr}\left(\mathbb{E}_q[\mathbf{\Lambda}^{k}_{n}]\mathrm{Var}\left(\mathbf{T}^{k}_{n}(i_n)\right)\right) +\mathbb{E}_q\left[\mathbf{T}^{k}_{n}(i_n)^{\top}\right]\mathbb{E}_q[\mathbf{\Lambda}^k_n]\mathbb{E}_q\left[\mathbf{T}^{k}_{n}(i_n)\right]\right\} +C\\
 & =\frac{\sum_nI_n}{2}\left(\sum_r^{R_{1,n}}\left(\psi(a_{1,n}^k)-\ln b_{1,n}^k\right) + \sum_r^{R_{n-1,n}}\left(\psi(a_{n-1,n}^k)-\ln b_{n-1,n}^k\right) + \sum_r^{R_{n,4}}\left(\psi(a_{n,4}^k)-\ln b_{n,4}^k\right)\right)\\
 &\quad-\frac{1}{2}\sum_{n=1}^{4}\mathrm{Tr}\left\{\mathbb{E}_q[\mathbf{\Lambda}^k_n]\left({\mathbf{T}^k_{n(n)}}^{\top}\mathbf{T}^k_{n(n)}\right)\right\} + C \\
& =\sum_{n_1=1}^{3}\sum_{n_2=2}^{4}\sum_{r}^{R_{n_1,n_2}}I_{n_1,n_2}\left(\psi(a_{n_1,n_2}^k)-\ln b_{n_1,n_2}^k\right) -\frac{1}{2}\sum_{n=1}^{4}\mathrm{Tr}\left\{\mathbb{E}_q[\mathbf{\Lambda}^k_n]\left({\mathbf{T}^k_{n(n)}}^{\top}\mathbf{T}^k_{n(n)}\right)\right\} + C\\
\end{aligned}
\end{equation*}	

\begin{equation*}
\begin{aligned}
\mathbb{E}_{q}&\left [   \sum_{n_1=1}^{3}\sum_{n_2=2}^{4}  \ln p\left(\boldsymbol\lambda^{k}_{n_1, n_2}\right)\right ]\\
& =\sum_{n_1=1}^{3}\sum_{n_2=2}^{4} \sum_{r}^{R_{n_1,n_2}}\left\{-\ln \Gamma\left(a_{0}^{k}\right)+a_{0}^{k} \ln d_{0}^{r}+\left(c_{0}^{k}-1\right) \mathbb{E}_{q}\left[\ln \boldsymbol\lambda^{k}_{n_1, n_2}\right]-d_{0}^{k} \mathbb{E}_{q}\left[\boldsymbol\lambda^{k}_{n_1, n_2}\right]\right\} \\
& =\sum_{n_1=1}^{3}\sum_{n_2=2}^{4} \sum_{r}^{R_{n_1,n_2}}\left\{-\ln \Gamma\left(a_{0}^{k}\right)+a_{0}^{k} \ln b_{0}^{k}+\left(a_{0}^{k}-1\right)\left(\psi\left(a_{{n_1, n_2}}^{k}\right)-\ln b_{n_1, n_2}^{k}\right)-b_{0}^{k} \frac{a_{{n_1, n_2}}^{k}}{b_{{n_1, n_2}}^{k}}\right\}\\
\end{aligned}
\end{equation*}	

\begin{equation*}
\begin{aligned}
\mathbb{E}_{q}[\ln p(\tau^k_h)] & =-\ln \Gamma\left(c_{0}^k\right)+c_{0}^k \ln d_{0}^k+\left(c_{0}^k-1\right) \mathbb{E}_{q}[\ln \tau]-d_{0}^k \mathbb{E}_{q}[\tau^k_h] \\
& =-\ln \Gamma\left(c_{0}^k\right)+c_{0}^k \ln d_{0}^k+\left(c_{0}^k-1\right)\left(\psi\left(c^k\right)-\ln d^k\right)-d_{0}^k \frac{c^k}{d^k}
\end{aligned}
\end{equation*}	

\begin{equation*}
\begin{aligned}
\mathbb{E}_{q}[\ln p(\tau^k_m)] & =-\ln \Gamma\left(e_{0}^k\right)+e_{0}^k \ln f_{0}^k+\left(e_{0}^k-1\right) \mathbb{E}_{q}[\ln \tau]-f_{0}^k \mathbb{E}_{q}[\tau^k_h] \\
& =-\ln \Gamma\left(e_{0}^k\right)+e_{0}^k \ln f_{0}^k+\left(e_{0}^k-1\right)\left(\psi\left(e^k\right)-\ln f^k\right)-f_{0}^k \frac{e^k}{f^k}
\end{aligned}
\end{equation*}	


\begin{equation*}
-\mathbb{E}_{q}\left[\sum_{n_1=1}^{3}\sum_{n_2=2}^{4}  \ln q\left(\boldsymbol\lambda^{k}_{n_1, n_2}\right)\right] =\sum_{n_1=1}^{3}\sum_{n_2=2}^{4}\sum_r^{R_{n_1,n_2}}\{\ln\Gamma(a_{n_1,n_2}^k)-(a_{n_1,n_2}^k-1)\psi(a_{n_1,n_2}^k)-\ln b_{n_1,n_2}^k+a_{n_1,n_2}^k\}                     
\end{equation*}	

\begin{equation*}
-\mathbb{E}_q\left[\ln q(\tau^k_h)\right]=\ln\Gamma(c^k)-(c^k-1)\psi(c^k)-\ln d^k+c^k
\end{equation*}	

\begin{equation*}
-\mathbb{E}_q\left[\ln q(\tau^k_m)\right]=\ln\Gamma(e^k)-(e^k-1)\psi(e^k)-\ln f^k+e^k
\end{equation*}	

In the above formula derivation, $\psi(\cdot)$ denotes the digamma function and $\Gamma(\cdot)$ denotes the Gamma function. In the Sec. \ref{sec:model}, we calculated the point estimation of $\mathcal{T}^{(k)}_n$. Therefore, in this part, we replace $\mathbb{E}_{q}\left[\mathcal{T}^{(k)}_n\right]$ with the point estimation of $\mathcal{T}^{(k)}_n$. Through analytical integration of all constituent terms and elimination of constant factors, we derive the following compact representation of the evidence lower bound ($ELBO$):

\begin{equation*}
\begin{aligned}
 ELBO & = -\frac{c^k}{2 d^k} \mathbb{E}_{q}\left[\left\|\mathcal {H}^{k}-\mathbb{FCTN}\left ( \mathcal{T}^{k}_1\times_1 \mathbf{P}_1,\mathcal{T}^{k}_2\times_2 \mathbf{P}_2,\mathcal{T}^{k}_3 ,\mathcal{T}^{k}_4\right)\right\|_{F}^{2}\right] \\
 &\quad -\frac{e^k}{2 f^k} \mathbb{E}_{q}\left[\left\|\mathcal {M}^{k}-\mathbb{FCTN}\left (\mathcal{T}^{k}_1,\mathcal{T}^{k}_2,\mathcal{T}^{k}_3 \times_3 \mathbf{P}_3,\mathcal{T}^{k}_4\right) \right\|_{\mathrm{F}}^2\right] \\
 &\quad-\frac{1}{2}\sum_{n=1}^{4}\mathrm{Tr}\left\{\mathbb{E}_q[\mathbf{\Lambda}^k_n]\left({\mathbf{T}^k_{n(n)}}^{\top}\mathbf{T}^k_{n(n)}\right)\right\} \\
 &\quad+\sum_{n_1=1}^{3}\sum_{n_2=2}^{4} \sum_{r}^{R_{n_1,n_2}}\left\{\ln\Gamma(a_{n_1,n_2}^k)+a_{n_1,n_2}^k\left(1-\ln b_{n_1,n_2}^k-\frac{b_0^k}{b_{n_1,n_2}^k}\right)\right\}\\
 &\quad+\ln\Gamma(c^k)+c^k(1-\ln d^k-\frac{d^k_0}{d^k})\\
 &\quad+\ln\Gamma(e^k)+e^k(1-\ln f^k-\frac{f^k_0}{f^k}) \\
 &\quad+ C
\end{aligned}
\end{equation*}	

\begin{figure*} [h!]
    \centering
    \includegraphics[width=\linewidth]{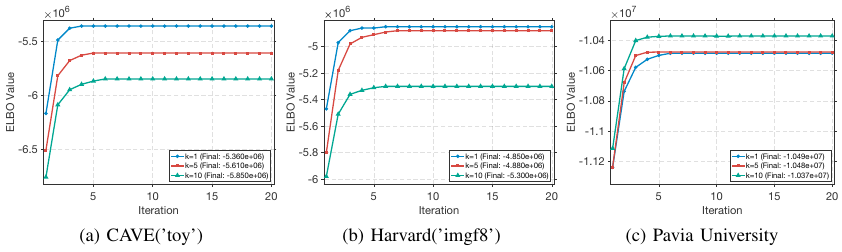}
    \caption{The convergence of $ELBO$ on three datasets.}
    \label{fig:ELBO} 
\end{figure*}

In order to empirically verify the theoretical monotonicity property of our $ELBO$ formulation, we performed a systematic convergence analysis on three benchmark datasets. It should be noted that the method is designed with group-independent inputs (each group is labeled as $k$), so the convergence processes of different groups of $k$ are independent of each other. Fig. \ref{fig:ELBO} illustrate the $ELBO$ trajectories under a consistent initialization scheme, in which all curves exhibit the guaranteed non-decreasing property, and the $ELBO$ increases to stability within five iterations, which satisfies the general criteria for convergence, and further demonstrates the stability and convergence of BFCTN.

\end{document}